\pdfoutput=1

\documentclass[11pt]{article}

\usepackage{acl}

\usepackage{times}
\usepackage{latexsym}

\usepackage[T1]{fontenc}

\usepackage[utf8]{inputenc}

\usepackage{microtype}
\usepackage[compact]{titlesec}

\usepackage{soul}
\usepackage{hyperref}
\usepackage{booktabs}
\usepackage{amsmath}
\usepackage{graphicx}
\usepackage{caption}
\usepackage{subcaption}
\usepackage{xcolor,colortbl}
\usepackage{siunitx}
\usepackage{tabularx}

\newcommand{\spacetf}{\hspace{1.2mm}}
\newcommand{\pst}{{ProSiT}}
\newcommand{\norm}[1]{\left\lVert#1\right\rVert}
\newcommand{\sota}{State-Of-The-Art}

\newcommand*\tablesize{%
  \@setfontsize\tablesize{11}{11}%
}
\makeatother

\title{\pst! Latent Variable Discovery with PROgressive SImilarity Thresholds}

\author{Tommaso Fornaciari \\
  Italian National Police
  \thanks{\hspace{2mm}\texttt{\href{mailto:tommaso.fornaciari@poliziadistato.it}{tommaso.fornaciari@poliziadistato.it}}} \\\And
  Dirk Hovy \\
  Bocconi University
  \thanks{\hspace{2mm}\texttt{\href{mailto:dirk.hovy@unibocconi.it}{dirk.hovy@unibocconi.it}}} \\\And
  Federico Bianchi\\
  Stanford University
  \thanks{\hspace{2mm}\texttt{\href{mailto:fede@stanford.edu}{fede@stanford.edu}}} \\
  }

\begin{document}
\maketitle
\begin{abstract}
The most common ways to explore latent document dimensions are topic models and clustering methods. However, topic models have several drawbacks: e.g., they require us to choose the number of latent dimensions \emph{a priori}, and the results are stochastic. Most clustering methods have the same issues and lack flexibility in various ways, such as not accounting for the influence of different topics on single documents, forcing word-descriptors to belong to a single topic (hard-clustering) or necessarily relying on word representations.
We propose PROgressive SImilarity Thresholds - \pst, a deterministic and interpretable method, agnostic to the input format, that finds the optimal number of latent dimensions and only has two hyper-parameters, which can be set efficiently via grid search.
We compare this method with a wide range of topic models and clustering methods on four benchmark data sets.
In most setting, \pst\ matches or outperforms the other methods in terms six metrics of topic coherence and distinctiveness, producing replicable, deterministic results.
\end{abstract}

\section{Introduction}
Latent variable models are essential for data exploration and are commonly used in social science and business applications. However, the most common methods used to explore these models, topic models and clustering, come with a set of drawbacks, chief among them their stochastic nature. This makes results hard to replicate precisely, as they require constant interpretation. 
In this paper, we instead propose a latent variable discovery method that produces deterministic results.

Topic models, such as latent Dirichlet allocation (LDA) \cite{blei2003latent}, rely on a number of topics that are provided \emph{a priori}, refining their initial random guesses to establish probability distribution iteratively.
This approach guarantees that the final state improves over the initial one in terms of topic quality, but the method is vulnerable to local optima.
This problem can be mitigated by initializing several models and picking the most useful/interpretable solution. Selection usually relies on coherence measures that estimate the topics' reliability.
However, the method also expects the user to provide the number of topics as a parameter.
Therefore the probability of finding good results depends on previous domain knowledge about the data set - which is often an unreasonable expectation.
In these cases, 
it is difficult to interpret or justify the resulting topics theoretically, as they depend on an arbitrary starting point.
Moreover, since each model's random initialization will provide slightly different results, these results are debatable.

Clustering methods, another popular choice for exploring latent document dimensions, similarly rely on the number of latent dimensions as input and their random initialization. 
For example, $K$-Means \cite{kmeans} can extract topics from groups of documents identified by minimizing the distance between the documents' representations.
The procedure partitions the vectors space in Voronoi cells, where each document exclusively contributes to the identification of a single cluster \cite{burrough2015principles}.
However, in many scenarios, the distinction between clusters/topics is nuanced, and we expect to find borderline texts that can be connected to different topics, which should not, in turn, be artificially forced to belong to a unique cluster.

To overcome these limitations, we propose \pst, a deterministic and interpretable latent variable discovery algorithm that is entirely reproducible.
\pst\ is agnostic to the input space (embeddings, discrete textual, and even non-textual features), allowing for flexible use and inheriting the properties of the input representations of whatever nature.
For example, with documents encoded by a multi-lingual model, the identified topics will also be multi-lingual.

\pst\ is entirely data-driven and does not depend on any guess or random initialization.
Instead, as the name suggests, it relies on the similarity between texts, i.e., their distances in the vector space.
The only theoretical assumption we make is that documents treating similar topic(s) will contain similar words, and therefore will be in proximate regions of the vector space.
In geometric terms, we assume the topic's convexity in the vector space, which implies that the distance between points represents their degree of similarity.

We evaluate \pst\ on four commonly used data sets against eight common and \sota\ latent variable discovery methods.
We evaluate the performance with six standard metrics of latent topic coherence and distinctiveness. We find that \pst\ is always comparable and often superior to \sota\ methods but requires fewer parameters and no prior selection of latent components.

\paragraph{Contributions}
We propose \pst, a novel algorithm for latent variable discovery. It is interpretable, deterministic, input agnostic, fuzzy, computationally efficient, and effective.
We release \pst\ as a PyPi package. The documentation can be found on GitHub.\footnote{\href{https://github.com/fornaciari/prosit}{https://github.com/fornaciari/prosit}}

\section{\pst}
\pst\ takes as input a corpus of documents. The algorithm has two main steps: 1) identifying the latent variables (from here on: topics) and 2) extracting words describing these topics. 

\pst\ is efficient for inductive learning: once trained, the topics are represented as points in the same vector space as the training documents.
To evaluate  unseen documents is as simple as computing their distance from the topics coordinates.
The only requirement for new documents is that they are represented in the same vector space as the training data.

\subsection{Step 1: Finding topics}
The first step is an iterative process that identifies several potential topic sets. The number of topics is data-driven, rather than decided \emph{a priori}.
\pst\ uses different degrees of similarity between groups of documents to determine whether they belong to a topic.
These similarity levels are determined by a threshold, which changes at each iteration step and follows a hyperbolic curve.
A parameter we call $\alpha$ tunes the slope of this curve.

Formally, we assume a set of documents $D$. Each document is represented in two ways: as a $d$-dimensional vector $m$ (continuous or sparse), and as a list of strings $s$, representing the (pre-processed) document with space-separated words.

We use the matrix $M$ over every document vector $m$ to determine the topics and $S$ over every string representation $s$ to extract the topic descriptors.

\begin{figure}
    \includegraphics[width=\linewidth]{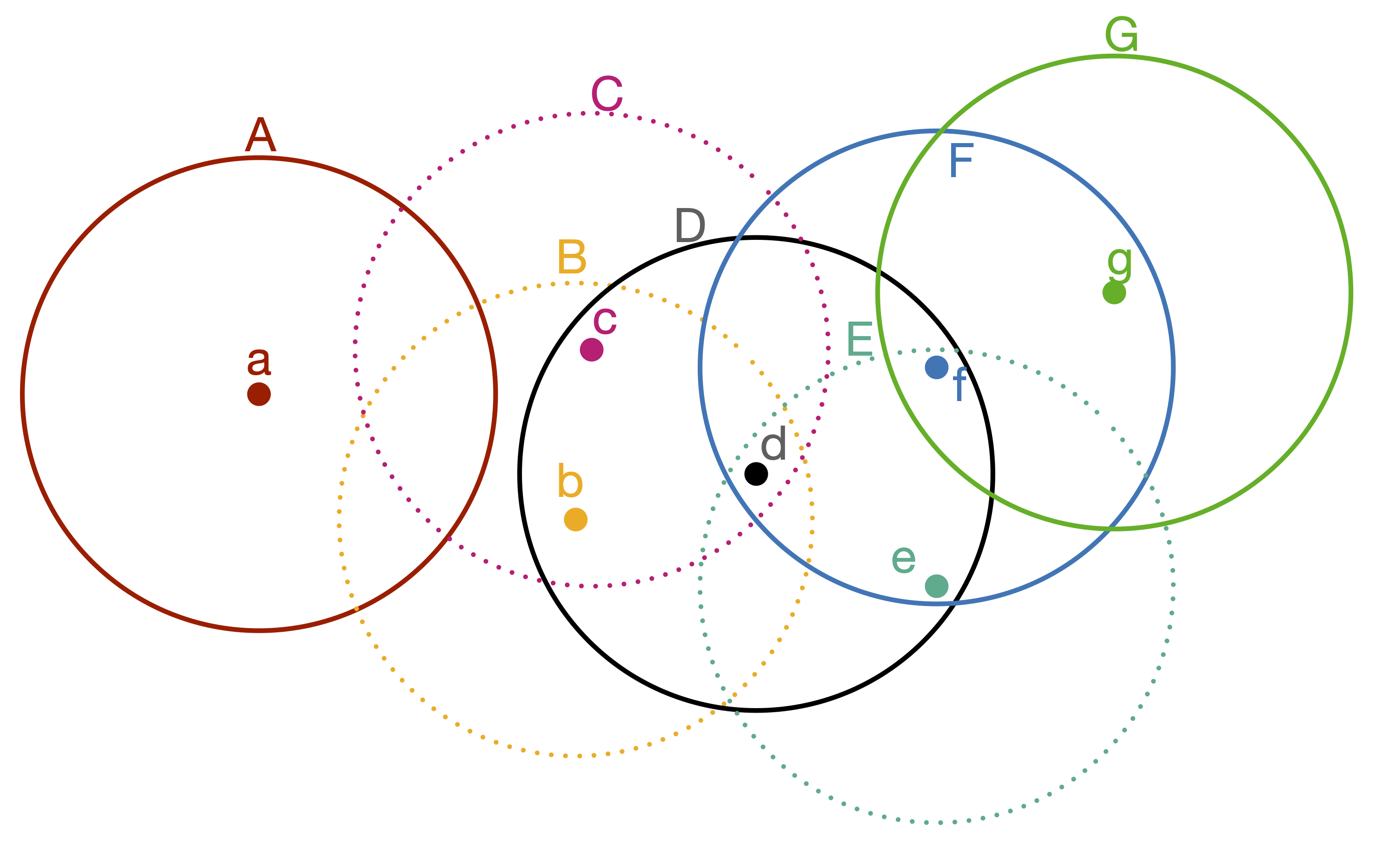}
    \caption{Examples of data points (lowercase letters) and the relative neighbourhood areas (uppercase letters), determined by $\alpha$. The dotted lined represent sets of neighbors that are subsets of other sets.}
    \label{fig:sets}
\end{figure}

We set a similarity threshold $\alpha$ and iterate over $M$ with the following procedure until it stops at the smallest number of topics found:
\begin{enumerate}
    \item removing any (possible) duplicated vectors to prevent topics being affected by repeated documents.
    This step could seem unnecessary, but many corpora contain repeated texts.
    Since \pst\ models topics as clusters of their associated documents' average representation,
    removing duplicates reduces the impact of repeated documents on topic identification.
    \item computing the cosine similarity between all documents in $M$:
        \begin{equation}
        C = \frac{M M^\top}{\norm{M} \norm{M^\top}}
        \end{equation}
    The norms are computed row-wise and multiplied as column and row vectors, respectively.
    The resulting matrix $C$ is a square, symmetric similarity matrix over the unique input documents. 
    \item mapping each data point to the set of all its neighbors (including itself), that is, the set of points more similar than the threshold $\alpha$.
    The output of this procedure is a list of redundant sets.
    However, it also identifies isolated instances too far from others, resulting in a singleton set.
    Both situations are simplified in Figure \ref{fig:sets}.
    
        \item removing the groups that are repeated subsets of other groups;
        \item computing the groups' centroids, i.e., the documents' average vector;
        \item assigning isolated instances (outliers) to the group having the closest centroid;
        \item recomputing the cluster centroids, including the outliers in the groups to which they were assigned.
    \item re-starting the iteration, using the centroids as new data points (keeping track of the original data points associated with these centroids)
\end{enumerate}

Similarly to agglomerative hierarchical clustering, each document is initially considered its own topic. The number of topics is reduced by iteratively collapsing the documents together.
However, \pst\ and agglomerative hierarchical clustering differ substantially. In the latter, pairs of instances are iteratively joined, creating a tree that is representable by dendrograms and in which each instance belongs to one cluster only.
In \pst, as shown in Figure \ref{fig:sets}, we create sets of instances lying within a similarity threshold, where overlap between sets is allowed. Therefore, a given instance can contribute to the centroid computation of more than one cluster, leading to smoother topic representations.

However, there is a problem: Iteratively averaging document vectors pushes the resulting centroids towards the mean of means, i.e., the global centroid of the entire data set.
To prevent the document representations from 
collapsing in one point,
in \pst\ we set an increasingly higher cosine similarity threshold ($CST$) at every training iteration $iter$ according to a hyperbolic function 
\begin{equation}
    CST = \frac{iter - \alpha}{iter}
\end{equation}
whose slope depends on the $\alpha$ parameter.

\begin{figure}
    \includegraphics[width=\linewidth]{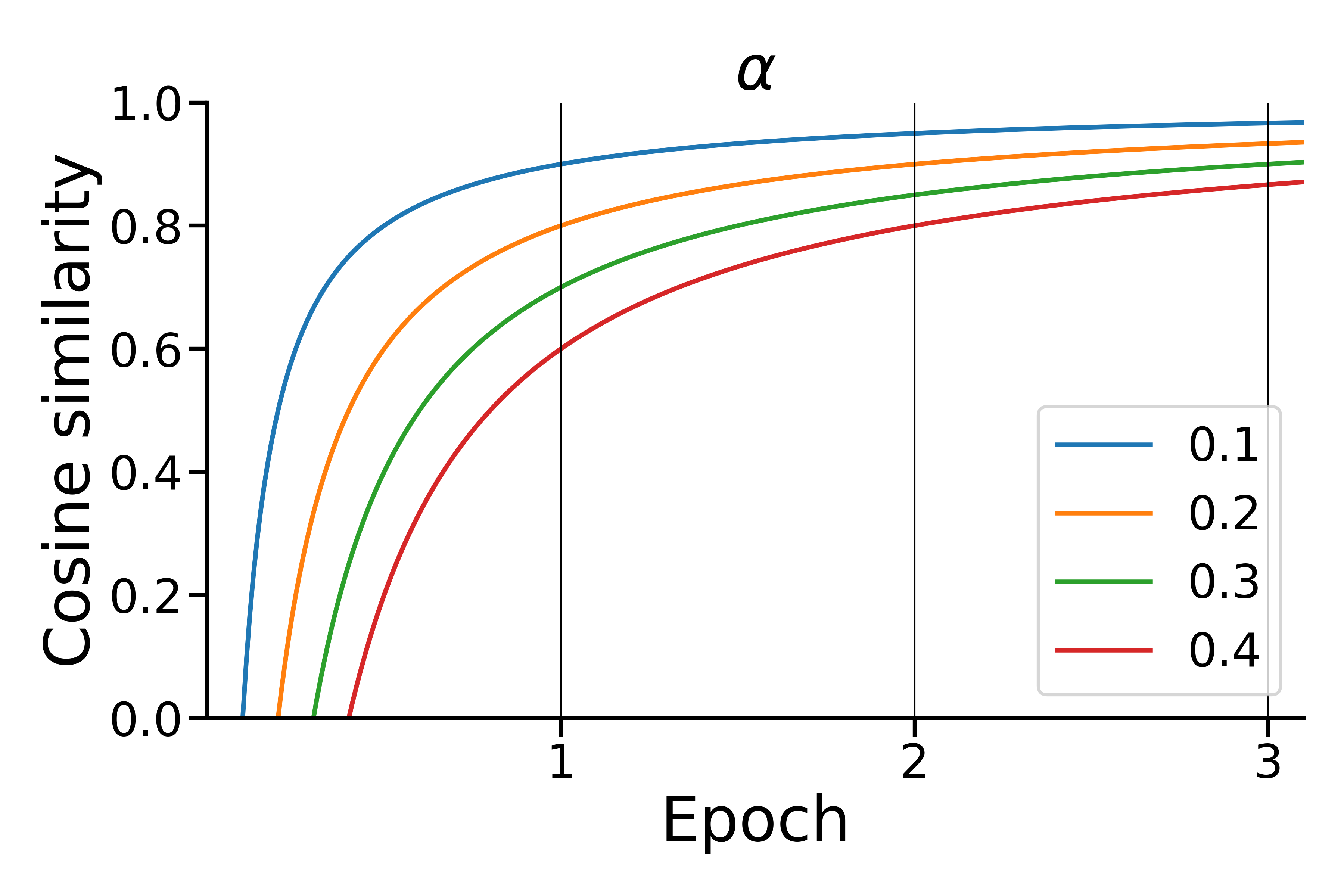}
    \caption{Some examples of the $\alpha$ curve.}
    \label{fig:alfa}
\end{figure}

Since the thresholds are cosine similarity values, the maximal value is 1. We will therefore want an $\alpha$ value close to 0. 
Figure \ref{fig:alfa} shows the hyperbolic curve for various values of $\alpha$.

\pst\ does not require the number of topics/clusters as an input: They emerge from the process.
Topics correspond to the number of clusters identified at each iteration, which become progressively lower until the algorithm reaches convergence.
The convergence is reached when all the computed centroids are farther (or less similar) from each other than the threshold, $\alpha$.

Since the topics are represented as points in the documents' vector space, we can compute the affinity between any (unseen) document and each topic in terms of distance, similarly to LDA, which measures the topics' presence in each document.

\subsection{Step 2: Selecting topic descriptors}
Once \pst\ has identified a set of topics, the second step consists of selecting the documents that best represent each topic and extracting the most representative terms from them.
While it is possible to extract the descriptors at every iteration (each corresponding to a different number of topics), for ease of exposition, we show the results for topics ranging between 5 and 25. 





In hard clustering methods, identifying the documents that provide the descriptors is straightforward, as every document belongs to a unique topic.
In our case, though, we consider the continuous distance between documents and topics so that documents can be associated with several topics with different distance levels.
We are interested in selecting the most representative documents, i.e., the closest ones to the topic centroids.
To achieve this, we use another threshold, $\beta$, to define how close to the topics a document should be for the topics to be used for descriptors.
$\beta$ describes $n$ percent documents closest to each topic centroid, ranging from 0 to 1 (where 1 means the whole data set).

Once we have selected the set of documents that are most representative of each topic, we extract the descriptors using the information gain (IG) \cite{forman2003extensive} value, computed for each topic.
IG measures the entropy of features selected from some instances belonging to different classes.
It is usually used for feature selection from instances in binary classification tasks.
Here we propose an original use of IG in a multi-class scenario, where the topics represent the classes.
We rank the terms according to their IG (i.e., probability of belonging to a topic) and select the top $n$ words.
This procedure, similarly to the document-topics' affinity, can measure the affinity between words and topics.

\section{Experimental Settings}

\subsection{Data sets}
\label{sec:data set}
We test \pst\ on four data sets, two with long documents and two with short documents. 
For long documents, we use the Reuters and Google News data sets,\footnote{The Reuters data set can be found at \url{https://www.nltk.org/book/ch02.html}.} which have previously been used by \newcite{sia-etal-2020-tired} and \newcite{qiang2020short}, respectively.
For short documents, we use Wikipedia abstracts from DBpedia,\footnote{The abstracts can be found at \url{https://wiki.dbpedia.org/ downloads-2016-10}.} the same data set used in by~\newcite{bianchi2020cross}, to which we refer as Wiki20K, and a tweet data set, used for topic models by \newcite{qiang2020short}.
Wiki20K~\cite{bianchi2020cross} contains Wikipedia abstracts filtered to consist of only the 2,000 most frequent words of the vocabulary.
Tweet and Google News are standard data sets in the community and were released by \newcite{qiang2020short}; both data sets have been preprocessed (e.g., stop words have been removed).
Table~\ref{tab:data set:descriptions} contains descriptive statistics for the data sets.
We use a small vocabulary size, which is desirable for most Topic Models scenarios, where including extremely-low frequency terms would result in a too fine-grained number of topics.

\begin{table}[t!]
    \centering
    \small
    \begin{tabular}{lrrrr} \toprule
         Data set & Docs & Vocab. & M. words & pre-p. \\
         \midrule
         Reuters & 10,788 & 949 & 130.11 & 55.02\\
         Google News & 10,950 & 2,000 & 191.98 & 68.00\\
         Wiki20K & 20,000 & 2,000 & 49.82 & 17.44\\
         Tweets2011 & 2,472 & 5098 & 8.56 & 8.56\\
\bottomrule
    \end{tabular}
    \caption{Corpora statistics, with vocabulary size and mean words with and without pre-processing.}
    \label{tab:data set:descriptions}
\end{table}

\subsection{Metrics}
\label{sub:metric}
We evaluate the topics with four metrics for coherence and two for distinctiveness.
First, we use standard coherence measures, that is $C_V$ and Normalized Pointwise Mutual Information (NPMI) \cite{roder2015exploring}.
Also, we consider the Rank-Biased Overlap (RBO) \cite{webber2010similarity}, a discrete measure of overlap between sequences. We also use the inverted RBO (IRBO) score, that is 1 - RBO \cite{bianchi2020pre}. This score describes how different the different topics are on average.
Lastly, similarly to the approach of \newcite{ding2018coherence}, we use an external word embedding-based coherence measure (WECO) to compute the coherence on an external domain. This metric computes the average pairwise similarity of the words in each topic and averages the results. We use the standard GoogleNews word embedding commonly used in the literature \cite{mikolov2013distributed}.

Concerning the distinctiveness, that is, how clearly the topics differentiate from each other, following \newcite{mimno2014low} we measure Topic Specificity (TS) and Topic Dissimilarity (TD). The first is the average Kullback-Leibler divergence \cite{kullback1951} from each topic’s conditional distribution to the corpus distribution; the second is based on the conditional distribution of words across different topics.

While we discuss the outcomes of all metrics, for space constraints, we only show the results of $C_V$, reporting the whole results in Appendix.



\subsection{Baselines}
\label{sub:base}
We compare \pst\ with two groups of models, differing by the text representations they require as input: embeddings or sparse count features.
\pst\ allows us to use each of them. 

All comparison methods require the number of latent topics as an \textit{a-priori} input parameter.
We evaluate their performance for inputs of 5, 10, 15, 20, and 25 topics to show a defined range. 
However, recall that \pst\ does not take the number of topics as input but instead identifies them automatically. 
Therefore, we also evaluate the other models on those numbers of topics that \pst\ finds.


\begin{table*}[ht!]
\centering
\footnotesize
\begin{tabular}{>{\columncolor[gray]{0.9}}l|>{\columncolor[gray]{0.85}}l>{\columncolor[gray]{0.9}}l>{\columncolor[gray]{0.9}}l|>{\columncolor[gray]{0.85}}l>{\columncolor[gray]{0.9}}l>{\columncolor[gray]{0.9}}l>{\columncolor[gray]{0.9}}l>{\columncolor[gray]{0.9}}l>{\columncolor[gray]{0.9}}l>{\columncolor[gray]{0.9}}l}
\toprule
\underline{Input}    &   \multicolumn{3}{>{\columncolor[gray]{0.9}}c|}{SBERT embeddings}   &    \multicolumn{7}{>{\columncolor[gray]{0.9}}c}{BoW}\\
Topics &         \pst &     CTM &    ZSTM &             \pst &    Agg. & Agg.+KM &      KM &  N-PLDA &     LDA &  DBScan \\
\midrule
5  &           0.7229* &   0.5697 &   0.5907 &  \textbf{0.7508}* &  0.5258 &  0.5133 &  0.5145 &   0.6438 &  0.5185 &         \\
6  &                   &   0.5626 &  0.5638* &           0.7387* &  0.5468 &  0.5548 &  0.5157 &   0.5337 &  0.5202 &         \\
7  &  \textbf{0.7617}* &   0.6299 &   0.5933 &                   &  0.5465 &  0.5316 &  0.5312 &  0.5997* &  0.5303 &         \\
8  &                   &  0.6795* &   0.6338 &           0.6643* &  0.5848 &  0.5601 &  0.5236 &    0.655 &  0.5061 &         \\
9  &           0.6941* &   0.6843 &   0.6833 &            0.6339 &  0.5784 &  0.5617 &  0.5628 &  0.6398* &  0.5069 &         \\
10 &                   &    0.667 &  0.7055* &                   &  0.5536 &  0.5528 &  0.5154 &  0.6436* &  0.5063 &         \\
11 &            0.6449 &   0.6701 &  0.6798* &                   &  0.5447 &  0.5462 &  0.5009 &  0.6456* &  0.5126 &         \\
13 &            0.6688 &   0.6302 &  0.6798* &            0.6461 &  0.5352 &  0.5257 &  0.5438 &  0.6516* &  0.5234 &         \\
15 &                   &   0.6698 &   0.692* &           0.6667* &  0.5403 &  0.5353 &  0.5342 &   0.6195 &  0.5256 &         \\
16 &                   &  0.6743* &   0.6647 &           0.6668* &  0.5269 &   0.535 &  0.5333 &   0.6515 &   0.522 &         \\
17 &            0.6384 &  0.6681* &   0.6291 &                   &  0.5388 &  0.5449 &  0.5431 &    0.64* &  0.5347 &         \\
20 &                   &   0.6312 &  0.6355* &                   &  0.5274 &  0.5315 &  0.5475 &  0.6387* &  0.5205 &  0.3726 \\
21 &            0.6244 &  0.6366* &   0.6348 &                   &  0.5404 &  0.5433 &   0.553 &  0.6171* &  0.5122 &         \\
25 &                   &   0.6424 &  0.6644* &                   &  0.5515 &   0.552 &  0.5716 &  0.6576* &  0.5202 &         \\
\bottomrule
\end{tabular}
\caption{Coherence values for Reuters data set. Bold: best group performance. *: best row performance. \pst\ columns differ by input formats.}
\label{tab:re_cv}
\end{table*}

In the first group, we consider contextualized topic models  \cite[CTM]{bianchi2020pre} and ZeroShot topic models \cite[ZSTM]{bianchi2020cross}.
They introduce the use of contextual embeddings in topic models.
Both rely on sentence-BERT \cite[SBERT]{reimers2019sentence} embeddings.

Second, we compare with models that take frequency-based bag-of-words (BoW) input representations.
These are agglomerative hierarchical cluster analysis (Agg.) \cite[Agg]{maimon2010data}, K-means (KM) \cite[KM]{kmeans}, K-means initialized with the agglomerative clustering centroids (Agg.+KM), Latent Dirichlet Allocation (LDA) \cite{blei2003latent}, neural-ProdLDA (N-PLDA) \cite[NPLDA]{srivastava2017autoencoding} and DBScan (DBScan) \cite{schubert2017dbscan}.

Neural-ProdLDA is a novel, state-of-the-art method, while LDA, agglomerative clustering and K-means are well-known and widely used techniques.
Similarly to \pst, DBScan discovers the number of topics as part of the training. Since the two methods tend to discover a different number of topics, a direct comparison is not possible. However, we show the DBScan performance for positioning purposes.

With the same constraint, we compare with Top2Vec \cite{angelov2020top2vec},  a recent method that jointly considers document and word representations.
We feed \pst\ with the same input (Doc2vec embeddings) and we show the performance, even though the number of topics differ.
Lastly, for positioning purposes, we show (for the two data sets considered in the released code) the performance of the procedure proposed by \newcite{sia-etal-2020-tired}, that relies on word representation and creates clusters using several algorithms
(Section \ref{app_sec_tired}).




\subsection{\pst\ tuning}
\pst's effectiveness depends on the choice of $\alpha$ and $\beta$.
We select them via grid search, evaluating the performance with the metrics mentioned above.
In this study, we found the optimal values ranged from $0.02$ to $\num{1e-06}$ for $\alpha$ and from $0.2$ to $0.03$ for $\beta$.
The small values for $\alpha$ here are not surprising, as the texts can be projected into very dense regions of the vector space.
In particular, \newcite{ethayarajh-2019-contextual} ``find that BERT embeddings occupy a
narrow cone in the vector space, and this effect increases from the earlier to later layers'', as is also pointed out by \newcite{rogers-etal-2020-primer}.  
We report the models with the best $C_V$, obtained through grid search. 
The other metrics are taken from those selected models. 

\begin{table*}[ht!]
\centering
\footnotesize
\begin{tabular}{>{\columncolor[gray]{0.9}}l|>{\columncolor[gray]{0.85}}l>{\columncolor[gray]{0.9}}l>{\columncolor[gray]{0.9}}l|>{\columncolor[gray]{0.85}}l>{\columncolor[gray]{0.9}}l>{\columncolor[gray]{0.9}}l>{\columncolor[gray]{0.9}}l>{\columncolor[gray]{0.9}}l>{\columncolor[gray]{0.9}}l>{\columncolor[gray]{0.9}}l}
\toprule
\underline{Input}    &   \multicolumn{3}{>{\columncolor[gray]{0.9}}c|}{SBERT embeddings}   &    \multicolumn{7}{>{\columncolor[gray]{0.9}}c}{BoW}\\
Topics &         \pst &     CTM &    ZSTM &             \pst &    Agg. & Agg.+KM &      KM &  N-PLDA &     LDA &  DBScan \\
\midrule
5  &          &    0.762 &           0.7831* &  0.8073 &  \textbf{0.8648}* &   0.8648 &   0.8648 &  0.7399 &  0.6805 &        \\
6  &   0.7454 &    0.723 &           0.7938* &  0.8047 &            0.858* &    0.858 &    0.858 &  0.7752 &  0.6529 &        \\
7  &    0.686 &     0.65 &  \textbf{0.8064}* &         &           0.8647* &   0.8647 &   0.8647 &  0.7369 &  0.6702 &        \\
10 &          &    0.704 &           0.7704* &         &            0.8317 &   0.8436 &  0.8524* &    0.72 &  0.6529 &        \\
11 &          &   0.6741 &           0.7122* &  0.7166 &            0.8368 &   0.8335 &  0.8402* &  0.6677 &  0.6594 &        \\
12 &  0.7503* &    0.688 &            0.6911 &         &           0.8484* &   0.8454 &   0.8294 &  0.7186 &   0.652 &        \\
13 &   0.7111 &   0.6952 &           0.7413* &   0.787 &            0.8557 &  0.8582* &   0.8508 &  0.7118 &  0.7152 &        \\
15 &          &   0.7057 &           0.7639* &         &            0.8587 &  0.8609* &   0.8587 &  0.7087 &  0.6983 &        \\
20 &   0.6445 &   0.7101 &           0.7137* &         &            0.8182 &   0.8168 &  0.8322* &  0.6672 &  0.6295 &        \\
22 &    0.667 &   0.6986 &            0.706* &  0.7372 &            0.8179 &   0.8167 &  0.8269* &  0.7264 &  0.6268 &        \\
25 &          &  0.6954* &            0.6813 &  0.7316 &            0.8202 &   0.8194 &  0.8267* &  0.6497 &  0.6644 &        \\
\bottomrule
\end{tabular}
\caption{Coherence values for Google News data set. Bold: best group performance. *: best row performance. \pst\ columns differ by input formats.}
\label{tab:gn_cv}
\end{table*}

\begin{table*}[ht!]
\centering
\footnotesize
\begin{tabular}{>{\columncolor[gray]{0.9}}l|>{\columncolor[gray]{0.85}}l>{\columncolor[gray]{0.9}}l>{\columncolor[gray]{0.9}}l|>{\columncolor[gray]{0.85}}l>{\columncolor[gray]{0.9}}l>{\columncolor[gray]{0.9}}l>{\columncolor[gray]{0.9}}l>{\columncolor[gray]{0.9}}l>{\columncolor[gray]{0.9}}l>{\columncolor[gray]{0.9}}l}
\toprule
\underline{Input}    &   \multicolumn{3}{>{\columncolor[gray]{0.9}}c|}{SBERT embeddings}   &    \multicolumn{7}{>{\columncolor[gray]{0.9}}c}{BoW}\\
Topics &         \pst &     CTM &    ZSTM &             \pst &    Agg. & Agg.+KM &      KM &  N-PLDA &     LDA &  DBScan \\
\midrule
5  &                   &  0.6109* &   0.6026 &  0.7339* &   0.6605 &  0.6812 &  0.6087 &            0.6241 &   0.552 &         \\
8  &           0.7223* &   0.6623 &    0.669 &  0.7623* &   0.7041 &  0.7154 &  0.6534 &            0.6896 &  0.5276 &         \\
10 &                   &  0.7301* &   0.6843 &          &  0.7243* &  0.7075 &  0.6744 &            0.7027 &  0.5537 &         \\
13 &  \textbf{0.7658}* &   0.7268 &   0.7295 &  0.7274* &   0.7272 &  0.7062 &  0.6969 &            0.7039 &  0.5596 &         \\
15 &            0.7308 &  0.7404* &   0.7246 &          &   0.7153 &  0.6859 &  0.7058 &           0.7627* &  0.5459 &         \\
18 &                   &    0.752 &  0.7538* &          &  0.7173* &  0.6946 &   0.704 &             0.702 &  0.5573 &  0.6302 \\
20 &                   &  0.7373* &   0.7314 &          &   0.7241 &  0.6999 &  0.6474 &           0.7592* &  0.5742 &         \\
23 &            0.7349 &     0.73 &   0.741* &  0.7604* &   0.7294 &  0.7153 &  0.6917 &            0.7214 &  0.5842 &         \\
24 &                   &   0.746* &   0.7339 &   0.752* &   0.7286 &  0.7179 &  0.6835 &            0.7284 &  0.6018 &         \\
25 &           0.7491* &   0.7448 &   0.7361 &          &   0.7303 &    0.72 &  0.6731 &  \textbf{0.7649}* &  0.5885 &         \\
\bottomrule
\end{tabular}
\caption{Coherence values for Wiki20K data set. Bold: best group performance. *: best row performance. \pst\ columns differ by input formats.}
\label{tab:db_cv}
\end{table*}

\begin{table*}[ht!]
\centering
\footnotesize
\begin{tabular}{>{\columncolor[gray]{0.9}}l|>{\columncolor[gray]{0.85}}l>{\columncolor[gray]{0.9}}l>{\columncolor[gray]{0.9}}l|>{\columncolor[gray]{0.85}}l>{\columncolor[gray]{0.9}}l>{\columncolor[gray]{0.9}}l>{\columncolor[gray]{0.9}}l>{\columncolor[gray]{0.9}}l>{\columncolor[gray]{0.9}}l>{\columncolor[gray]{0.9}}l}
\toprule
\underline{Input}    &   \multicolumn{3}{>{\columncolor[gray]{0.9}}c|}{SBERT embeddings}   &    \multicolumn{7}{>{\columncolor[gray]{0.9}}c}{BoW}\\
Topics &         \pst &     CTM &    ZSTM &             \pst &    Agg. & Agg.+KM &      KM &  N-PLDA &     LDA &  DBScan \\
\midrule
5  &                   &  0.4553* &   0.3935 &                   &    0.702 &  0.7022* &    0.585 &  0.3761 &  0.4372 &         \\
6  &                   &  0.4115* &   0.3929 &  \textbf{0.7153}* &   0.6622 &   0.6533 &   0.6809 &  0.3811 &  0.3966 &         \\
10 &  \textbf{0.5909}* &   0.4242 &   0.4086 &                   &  0.6675* &   0.6472 &   0.6395 &   0.398 &  0.4119 &         \\
11 &                   &   0.4164 &  0.5114* &            0.5132 &  0.6447* &   0.6304 &   0.6309 &  0.4018 &   0.397 &         \\
15 &                   &   0.4481 &  0.4593* &                   &  0.6584* &   0.6536 &   0.6379 &  0.4451 &  0.3887 &         \\
19 &           0.5788* &   0.4871 &   0.4582 &                   &   0.6541 &  0.6604* &   0.6321 &  0.4135 &   0.403 &         \\
20 &                   &  0.5151* &   0.4823 &                   &   0.6594 &   0.6621 &  0.6896* &   0.464 &  0.3978 &         \\
23 &                   &  0.4859* &   0.4594 &                   &  0.6732* &   0.6706 &   0.6522 &   0.425 &   0.428 &  0.5193 \\
25 &                   &  0.4782* &   0.4499 &                   &   0.6786 &   0.6788 &  0.6859* &   0.519 &  0.4023 &         \\
\bottomrule
\end{tabular}
\caption{Coherence values for Tweet data set. Bold: best group performance. *: best row performance. \pst\ columns differ by input formats.}
\label{tab:tw_cv}
\end{table*}

\section{Results}
Tables \ref{tab:re_cv} to \ref{tab:tw_cv} show the coherence values for the four data sets.
Vertical lines separate different sets of results, grouped according to the kind of input provided to the models, as pointed out in Section \ref{sub:base}.
\pst's outcomes are reported in the darkest columns and compared with the columns to their right.
The results in bold underline the highest performance in each group.
The asterisk indicates the overall highest row performance, i.e., by topic number.

In three data sets, Reuters, Wiki20K, and Tweet, \pst's $C_V$ mostly outperforms the baselines. 
On Google News, \pst's performance is lower but still near the highest coherence scores for embeddings and count-based inputs and outperforms N-PLDA and LDA.
The NPMI score mostly reproduces the same pattern. 
\pst\ also shows high values of IRBO, which measures the topics' diversity.
This outcome is noticeable, as \pst\ is not a hard-clustering method, i.e., it allows the use of the same words in different topics.
Finally, the WECO score, which measures the out-of-domain coherence, shows more mixed results: \pst\ obtains results higher than the baselines in the Reuters and Twitter data sets but not in Wiki20K and Google News, where the results are not the best, but still close to those of the baselines.

\begin{table*}[ht!]
\centering
\footnotesize
\begin{tabular}{ll}
\toprule
Topic & Descriptors\\
\midrule
1 	 & egypt,\spacetf journalist,\spacetf detained,\spacetf attacked,\spacetf targeted,\spacetf beaten,\spacetf list,\spacetf mubarak,\spacetf arrested,\spacetf cairo\\
2 	 & sundance,\spacetf festival,\spacetf film,\spacetf celebrity,\spacetf rite,\spacetf movie,\spacetf photo,\spacetf sighting,\spacetf medium,\spacetf review\\
3 	 & speech,\spacetf king,\spacetf award,\spacetf oscar,\spacetf guild,\spacetf win,\spacetf nomination,\spacetf sag,\spacetf actor,\spacetf top\\
4 	 & fishing,\spacetf fly,\spacetf fish,\spacetf bass,\spacetf caught,\spacetf trout,\spacetf tip,\spacetf gear,\spacetf steelhead,\spacetf salmon\\
5 	 & bowl,\spacetf super,\spacetf aguilera,\spacetf christina,\spacetf anthem,\spacetf national,\spacetf video,\spacetf xlv,\spacetf eminem,\spacetf volkswagen\\
6 	 & acai,\spacetf berry,\spacetf weight,\spacetf loss,\spacetf diet,\spacetf plan,\spacetf healthy,\spacetf reduction,\spacetf benefit,\spacetf lose\\
7 	 & commercial,\spacetf superbowl,\spacetf ad,\spacetf doritos,\spacetf favorite,\spacetf youtube,\spacetf chrysler,\spacetf pepsi,\spacetf beetle,\spacetf vw\\
8 	 & fda,\spacetf approves,\spacetf birth,\spacetf drug,\spacetf preterm,\spacetf news,\spacetf treat,\spacetf risk,\spacetf reduce,\spacetf depression\\
9 	 & cpj,\spacetf website,\spacetf protest,\spacetf blocked,\spacetf amid,\spacetf egypt,\spacetf journalist,\spacetf judge,\spacetf law,\spacetf strike\\
10   & judge,\spacetf law,\spacetf health,\spacetf care,\spacetf federal,\spacetf rule,\spacetf obama,\spacetf reform,\spacetf strike,\spacetf ruled\\
\bottomrule
\end{tabular}
\caption{Descriptors for 10 topics from Tweet data set.}
\label{tab:desc}
\end{table*}

Concerning the Dissimilarity (TD), \pst\ outperforms the other methods in most conditions and data sets. The specificity (TS) still shows a less pronounced \pst's prevalence, even though it still outperforms other methods in many conditions.
This outcome is expected, as the IG used for the descriptors' selection minimizes' the terms entropy, enhancing TD, while no hard-clustering constraints are required, which would emphasize TS.
The results on NPMI, IRBO, WECO, TS, and TD are shown in Appendix.

\begin{figure}
    \includegraphics[width=\linewidth]{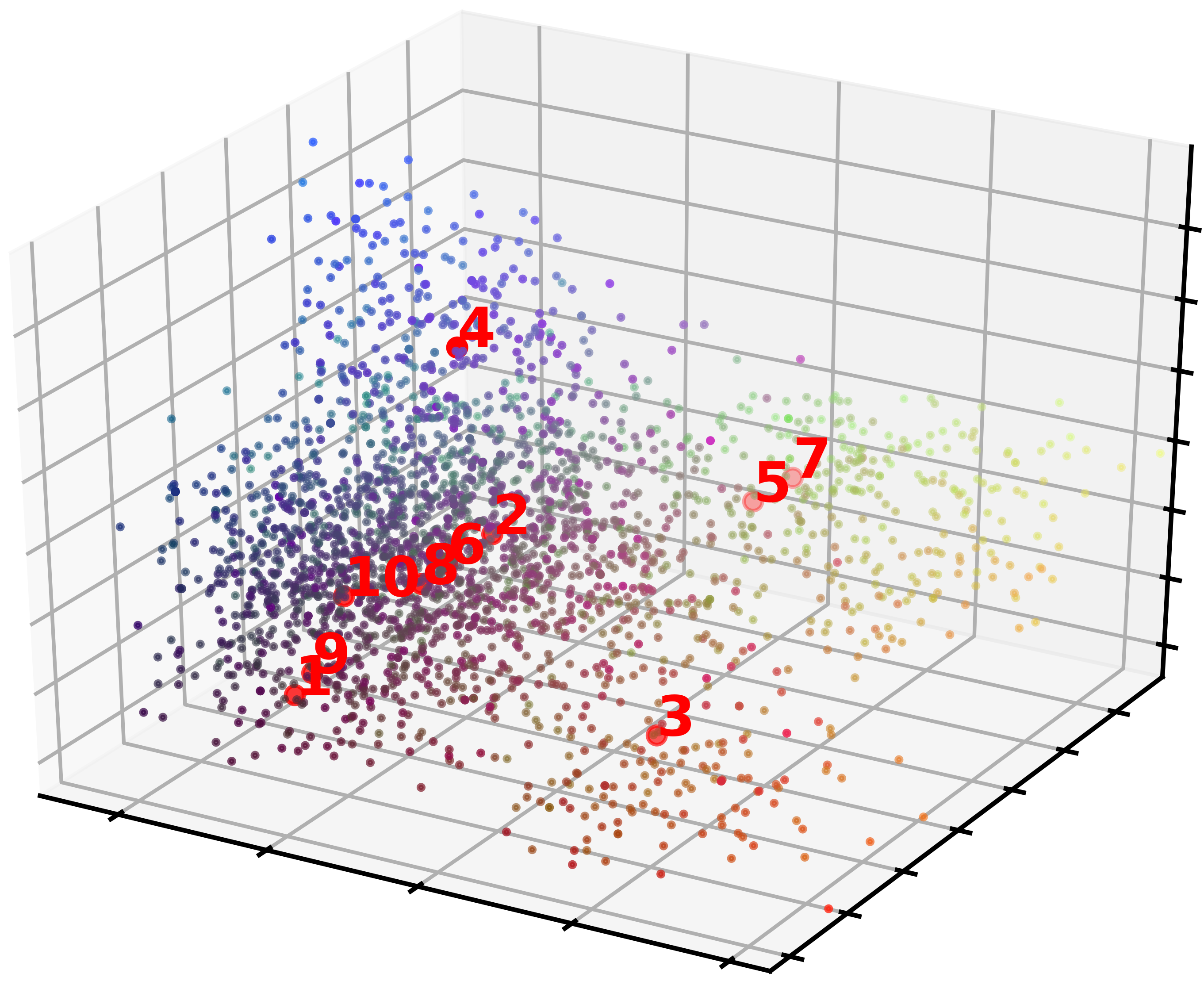}
    \caption{Text points and topics' from Table \ref{tab:desc}} 
    \label{fig:svd}
\end{figure}

To give a concrete sense of the topics identified with \pst, in Table \ref{tab:desc} we list the descriptors identified for Tweet with 
$\alpha = 0.0001, \beta = 0.2$ 
and SBERT representations as input. The associated performance in Table \ref{tab:tw_cv}, first column, 10-topics is $C_V = 0.5909$. 
Figure \ref{fig:svd} shows the same topics with the text data points.
Their dimensionality was reduced with truncated Singular Value Decomposition (SVD) \cite{halko2011finding}.

\section{Discussion}
According to various metrics, the results show that \pst\ can extract from corpora latent topics comparable to or even better than several standard and state-of-the-art models.

Google News is the only data set for which \pst\ does not exceed the other methods. 
The most effective models are agglomerative clustering and K-means, which outperform not only \pst\ but also neural-ProdLDA, LDA, and the CTMs (even though, in the last two cases, the models take different inputs).

This dominance of clustering methods is indicative. Agglomerative clustering follows a hierarchical procedure to separate the documents, minimizing the distance between data points; similarly, K-means minimizes the clusters' inertia.
Their objectives make these models most effective when the clusters are convex and isotropic.
The results suggest that this is the case with Google News.
Conversely, during the training iterations, \pst's centroids have some freedom to move away from their original position.
The centroid position is also affected by outliers, which makes \pst\ suitable for non-isotropic spaces; this explains \pst's versatility on the other data sets. 
This intuition is confirmed by the IRBO values, where \pst\ outperforms the other models on Google News. This result indicates its ability to create diverse topics, incorporating peripheral data points more effectively in the clusters than agglomerative clustering and K-means.

\pst\ also offers vast possibilities for results interpretation.
Since documents and topics are represented as points in the same vector space, the relation of each document with the topics can be expressed in terms of distance.
Passing those distances to a SoftMax function \cite{goodfellow2016deep}, we obtain a probability distribution that describes the affinity of each document with each topic; this is similar to the $\theta$ distribution in LDA.
Analogously, the IG gives the probability of each word belonging to the different topics. This distribution provides a value for every word in the vocabulary, not just the descriptors.

\section{Related Work}
\label{sec:relwork}
This work proposes a possible solution to the drawbacks of the most common Topic Models algorithms.
Aimed to overcome the necessity of guessing the topics' number, \newcite{broderick2013mad} rely on a Bayesian non-parametric framework that generates priors for the topics' identification.
Similar goal characterizes the family of Spectral Topic Models \cite{anandkumar2012method,arora2012learning,mimno2014low,lee2020prior}.
However, differently from \pst, they still rely on some stochastic process to obtain the topics' distributions. 

The success of deep learning methods in NLP has also fostered new methods in topic models.
The most recent ones are the contextualized topic models (CTM), which have been proposed in two studies \cite{bianchi2020cross,bianchi2020pre}.
They represent the first approach to incorporate the semantic knowledge of pre-trained language models like BERT \cite{devlin:bert} into topic models.
Exploiting BERT's multi-lingual models, CTMs can map documents from different languages in a unified space.
CTMs build on the neural-ProdLDA \cite{srivastava2017autoencoding}, a neural topic model based on variational autoencoders.
These methods create latent document representations from which they reconstruct the documents' words, approximating the Dirichlet prior with Gaussian distributions.
\pst\ does not use Dirichlet priors.
Also, CTMs and neural-ProdLDA require the number of topics as an input parameter, which we prefer to avoid as we believe this number should evolve from the data analysis.
However, similarly to CTM, \pst\ can use pre-trained language model representations as input, with the subsequent possibility of building multi-lingual topic models.

Deep learning methods are also the base for Top2Vec by \newcite{angelov2020top2vec}.
Similarly to \pst, this method identifies topics in the same vector space as the documents, aggregating them according to their similarity. 
However, the author determines the topics' descriptors from the position of word embeddings, which have to lie in the same vector space.
Conversely, \pst\ extracts the topic descriptors directly from the documents.
This feature makes \pst\ more flexible concerning different document representations.
\newcite{sia-etal-2020-tired} proposed a clustering procedure that creates topics starting from word embeddings, which are re-ranked using document information.
Their procedure performs well compared to an LDA baseline (which \pst\ also easily beats). Recently, \newcite{grootendorst2022bertopic} has proposed BERTopic, an effective model that first clusters documents together using sentence embeddings and then selects the most relevant keywords using class-based TF-IDF. We leave the comparison of ProSiT and BERTopic for future work.

\newcite{gialampoukidis2016hybrid} propose a hybrid procedure, which relies on DBSCAN \cite{ester1996density} to determine the clusters, whose number is used as the number of topics to compute LDA. 
This procedure is an effective way to provide LDA with a qualified prior on the number of topics.
Indeed, DBSCAN is a widely used method for topic models \cite{schubert2017dbscan}.
Using the points' density, it can determine an optimal number of topics.
However, it is not entirely deterministic (it starts from random points, and the cluster boundaries can be identified in different ways).
Also, DBSCAN discards outliers, which is not a desirable feature for topic models, where \textit{every} document should be evaluated. 
Its computational efficiency was a subject of debate, with differing opinions among researchers \cite{gan2015dbscan,schubert2017dbscan}.
\pst, in contrast, uses all documents.
The previous literature focuses on well-known methods for topic models and clustering, such as LDA \cite{blei2003latent} and K-means \cite{kmeans}.
Our method is more similar to the latter, as it derives topics elaborating from the original position of the documents in the vector space.



\section{Conclusions}
Latent topic models are crucial tools for data exploration.
However, in many scientific fields, such as the social sciences, and many application areas, such as legal contexts, the theoretical motivation of the analyses is as essential as the quality of their outcome.
The random initialization and necessary \emph{a priori} decision of the number of topics are particularly weak motivations and can invalidate even high-quality topics unusable for those areas.
\pst\ provides a viable answer to the need for interpretable high-quality topic models that are data driven rather than imposed \emph{a priori}.
\pst's outcomes are transparent: they reflect the degrees of similarity explored to cluster the documents (for the topics' identification) and the degrees of proximity between documents and topics (for the topic descriptors' extraction).
This is preferable to guess directly the topics' number, as the outcomes lie on a meaningful continuum, rather than resulting from a (random) guess

\pst's performance is comparable or higher than that of several other state-of-the-art methods.
\pst's setup allows users to explore several texts' representations, from dense embeddings to sparse feature vectors.


\section*{Ethical considerations}
\pst\ is a method to cluster documents according to their latent similarity; we do not consider this procedure harmful \emph{per se}.
However, as the input documents and their representations can carry unwanted biases, unethical content, and personal information, this could be reflected in \pst's outcomes.
Therefore, we invite a careful and responsible use of this method.

\bibliographystyle{acl_natbib}
\bibliography{custom}

\appendix

\begin{table*}[ht]
\section{Appendix} 

\subsection{NPMI values}
\label{app_sec_nmpi}


\centering
\tiny
\begin{tabular}{>{\columncolor[gray]{0.9}}l|>{\columncolor[gray]{0.85}}l>{\columncolor[gray]{0.9}}l>{\columncolor[gray]{0.9}}l|>{\columncolor[gray]{0.85}}l>{\columncolor[gray]{0.9}}l>{\columncolor[gray]{0.9}}l>{\columncolor[gray]{0.9}}l>{\columncolor[gray]{0.9}}l>{\columncolor[gray]{0.9}}l>{\columncolor[gray]{0.9}}l}
\toprule
\underline{Input}    &   \multicolumn{3}{>{\columncolor[gray]{0.9}}c|}{SBERT embeddings}   &    \multicolumn{7}{>{\columncolor[gray]{0.9}}c}{BoW}\\
Topics &         \pst &     CTM &    ZSTM &             \pst &    Agg. & Agg.+KM &      KM &  N-PLDA &     LDA &  DBScan \\
\midrule
5  &  0.1749* &   0.0672 &            0.0577 &           0.1832* &  0.0799 &  0.0713 &  0.0741 &   0.0363 &   0.074 &          \\
6  &          &  0.0667* &            0.0588 &  \textbf{0.2124}* &  0.0889 &  0.1071 &  0.0755 &   0.0215 &  0.0686 &          \\
7  &  0.1653* &   0.0853 &             0.104 &                   &  0.0826 &  0.0907 &  0.0902 &  0.1056* &   0.069 &          \\
8  &          &  0.1405* &             0.125 &           0.1416* &  0.1112 &  0.1103 &  0.0888 &   0.1225 &  0.0581 &          \\
9  &    0.113 &  0.1463* &            0.1381 &             0.142 &  0.1057 &  0.1055 &  0.1098 &  0.1494* &  0.0581 &          \\
10 &          &   0.1652 &  \textbf{0.2032}* &                   &  0.0972 &  0.1041 &  0.0795 &  0.1593* &  0.0599 &          \\
\bottomrule
\end{tabular}
\caption{NPMI values for Reuters data set.}
\label{tab:re_npmi}

\begin{tabular}{>{\columncolor[gray]{0.9}}l|>{\columncolor[gray]{0.85}}l>{\columncolor[gray]{0.9}}l>{\columncolor[gray]{0.9}}l|>{\columncolor[gray]{0.85}}l>{\columncolor[gray]{0.9}}l>{\columncolor[gray]{0.9}}l>{\columncolor[gray]{0.9}}l>{\columncolor[gray]{0.9}}l>{\columncolor[gray]{0.9}}l>{\columncolor[gray]{0.9}}l}
\toprule
\underline{Input}    &   \multicolumn{3}{>{\columncolor[gray]{0.9}}c|}{SBERT embeddings}   &    \multicolumn{7}{>{\columncolor[gray]{0.9}}c}{BoW}\\
Topics &         \pst &     CTM &    ZSTM &             \pst &    Agg. & Agg.+KM &      KM &  N-PLDA &     LDA &  DBScan \\
\midrule
5  &          &   0.1947 &           0.2058* &  0.1793 &  0.2975* &   0.2975 &            0.2975 &  0.1364 &   0.076 &        \\
6  &   0.1247 &   0.1453 &           0.1989* &  0.2098 &   0.285* &    0.285 &             0.285 &  0.1822 &  0.0674 &        \\
7  &   0.1455 &   0.0811 &  \textbf{0.2076}* &         &  0.2956* &   0.2956 &            0.2956 &  0.1489 &  0.0798 &        \\
10 &          &   0.1317 &           0.1806* &         &   0.2768 &   0.2813 &            0.286* &  0.1101 &  0.0699 &        \\
11 &          &   0.1021 &           0.1189* &  0.1195 &  0.2837* &   0.2829 &            0.2832 &  0.0784 &  0.0712 &        \\
12 &  0.1761* &   0.1052 &            0.1285 &         &  0.3012* &   0.3004 &            0.2605 &  0.1306 &  0.0668 &        \\
\bottomrule
\end{tabular}
\caption{NPMI values for Google News data set.}
\label{tab:gn_npmi}

\begin{tabular}{>{\columncolor[gray]{0.9}}l|>{\columncolor[gray]{0.85}}l>{\columncolor[gray]{0.9}}l>{\columncolor[gray]{0.9}}l|>{\columncolor[gray]{0.85}}l>{\columncolor[gray]{0.9}}l>{\columncolor[gray]{0.9}}l>{\columncolor[gray]{0.9}}l>{\columncolor[gray]{0.9}}l>{\columncolor[gray]{0.9}}l>{\columncolor[gray]{0.9}}l}
\toprule
\underline{Input}    &   \multicolumn{3}{>{\columncolor[gray]{0.9}}c|}{SBERT embeddings}   &    \multicolumn{7}{>{\columncolor[gray]{0.9}}c}{BoW}\\
Topics &         \pst &     CTM &    ZSTM &             \pst &    Agg. & Agg.+KM &      KM &  N-PLDA &     LDA &  DBScan \\
\midrule
5  &                   &  0.0148* &    0.005 &            0.204* &   0.1364 &  0.1447 &   0.1129 &     0.02 &   0.087 &        \\
8  &           0.2003* &   0.1125 &   0.1282 &  \textbf{0.2227}* &   0.1661 &  0.1743 &   0.1282 &   0.1151 &   0.071 &        \\
10 &                   &    0.133 &  0.1472* &                   &  0.1867* &  0.1827 &   0.1612 &   0.1762 &  0.0751 &        \\
13 &  \textbf{0.2218}* &   0.1419 &   0.1656 &           0.2052* &   0.1854 &  0.1762 &    0.177 &   0.1328 &  0.0833 &        \\
15 &           0.1913* &   0.1592 &   0.1687 &                   &   0.1766 &  0.1661 &   0.1751 &  0.1874* &  0.0784 &        \\
18 &                   &  0.1809* &   0.1781 &                   &   0.1772 &  0.1677 &  0.1804* &   0.1602 &  0.0795 &  0.066 \\
20 &                   &   0.1633 &  0.1783* &                   &   0.1805 &  0.1704 &   0.1449 &  0.1945* &  0.0992 &        \\
23 &            0.1917 &   0.1862 &  0.2054* &           0.2004* &   0.1846 &   0.179 &   0.1609 &   0.1787 &  0.1026 &        \\
24 &                   &   0.1861 &  0.1954* &           0.2159* &   0.1861 &  0.1817 &   0.1602 &   0.1897 &  0.1124 &        \\
25 &           0.1984* &   0.1938 &   0.1829 &                   &   0.1853 &  0.1809 &   0.1495 &  0.2008* &  0.1019 &        \\
\bottomrule
\end{tabular}
\caption{NPMI values for Wiki20K data set.}
\label{tab:db_npmi}

\begin{tabular}{>{\columncolor[gray]{0.9}}l|>{\columncolor[gray]{0.85}}l>{\columncolor[gray]{0.9}}l>{\columncolor[gray]{0.9}}l|>{\columncolor[gray]{0.85}}l>{\columncolor[gray]{0.9}}l>{\columncolor[gray]{0.9}}l>{\columncolor[gray]{0.9}}l>{\columncolor[gray]{0.9}}l>{\columncolor[gray]{0.9}}l>{\columncolor[gray]{0.9}}l}
\toprule
\underline{Input}    &   \multicolumn{3}{>{\columncolor[gray]{0.9}}c|}{SBERT embeddings}   &    \multicolumn{7}{>{\columncolor[gray]{0.9}}c}{BoW}\\
Topics &         \pst &     CTM &    ZSTM &             \pst &    Agg. & Agg.+KM &      KM &  N-PLDA &     LDA &  DBScan \\
\midrule
5  &                   &  -0.2336 &  -0.1261* &                   &  0.1733 &  0.1734* &   0.1366 &  -0.2384 &  -0.3048 &         \\
6  &                   &  -0.2005 &   -0.145* &  \textbf{0.3094}* &   0.112 &   0.1256 &   0.1632 &  -0.2621 &   -0.191 &         \\
10 &  \textbf{0.1784}* &  -0.0997 &   -0.0981 &                   &   0.191 &  0.1916* &   0.1786 &  -0.0612 &    -0.23 &         \\
11 &                   &  -0.1069 &  -0.0259* &            0.1144 &  0.1865 &   0.1804 &  0.2092* &  -0.1405 &   -0.213 &         \\
15 &                   &  -0.0266 &   0.0219* &                   &  0.2145 &  0.2175* &   0.1929 &  -0.0406 &   -0.191 &         \\
19 &           0.1522* &   0.0115 &    0.0006 &                   &  0.2161 &  0.2217* &   0.2211 &  -0.0713 &  -0.1456 &         \\
20 &                   &   0.0331 &   0.0578* &                   &  0.2263 &   0.2333 &  0.2592* &   0.0421 &  -0.1677 &         \\
23 &                   &  0.0429* &     0.028 &                   &  0.2507 &  0.2546* &   0.2437 &   0.0442 &  -0.1689 &  0.0543 \\
25 &                   &  0.0331* &    0.0078 &                   &  0.2602 &  0.2622* &   0.2572 &   0.0998 &  -0.1571 &         \\
\bottomrule
\end{tabular}
\caption{NPMI values for Tweet data set.}
\label{tab:tw_npmi}
\end{table*}

\begin{table*}[htpb]
\subsection{IRBO values}
\label{app_sec_rbo}


\centering
\tiny
\begin{tabular}{>{\columncolor[gray]{0.9}}l|>{\columncolor[gray]{0.85}}l>{\columncolor[gray]{0.9}}l>{\columncolor[gray]{0.9}}l|>{\columncolor[gray]{0.85}}l>{\columncolor[gray]{0.9}}l>{\columncolor[gray]{0.9}}l>{\columncolor[gray]{0.9}}l>{\columncolor[gray]{0.9}}l>{\columncolor[gray]{0.9}}l>{\columncolor[gray]{0.9}}l}
\toprule
\underline{Input}    &   \multicolumn{3}{>{\columncolor[gray]{0.9}}c|}{SBERT embeddings}   &    \multicolumn{7}{>{\columncolor[gray]{0.9}}c}{BoW}\\
Topics &         \pst &     CTM &    ZSTM &             \pst &    Agg. & Agg.+KM &      KM &  N-PLDA &     LDA &  DBScan \\
\midrule
5  &   0.9566 &  \textbf{1.0}* &      1.0 &  \textbf{1.0}* &  0.6242 &  0.6681 &  0.6743 &     1.0* &  0.7094 &         \\
6  &          &         0.9852 &  0.9974* &         0.9912 &  0.6598 &  0.6673 &  0.5855 &     1.0* &   0.727 &         \\
7  &   0.9757 &         0.9899 &  0.9927* &                &  0.6683 &  0.6433 &  0.6386 &     1.0* &  0.7406 &         \\
8  &          &         0.9873 &  0.9968* &           1.0* &  0.6756 &  0.6485 &  0.6474 &   0.9942 &  0.7385 &         \\
9  &  0.9931* &         0.9785 &    0.984 &           1.0* &  0.7008 &  0.6697 &  0.6444 &   0.9903 &  0.7589 &         \\
10 &          &         0.9796 &  0.9869* &                &  0.6967 &  0.6656 &  0.6532 &  0.9816* &  0.7588 &         \\
\bottomrule
\end{tabular}
\caption{IRBO values for Reuters data set.}
\label{tab:re_rbo}

\begin{tabular}{>{\columncolor[gray]{0.9}}l|>{\columncolor[gray]{0.85}}l>{\columncolor[gray]{0.9}}l>{\columncolor[gray]{0.9}}l|>{\columncolor[gray]{0.85}}l>{\columncolor[gray]{0.9}}l>{\columncolor[gray]{0.9}}l>{\columncolor[gray]{0.9}}l>{\columncolor[gray]{0.9}}l>{\columncolor[gray]{0.9}}l>{\columncolor[gray]{0.9}}l}
\toprule
\underline{Input}    &   \multicolumn{3}{>{\columncolor[gray]{0.9}}c|}{SBERT embeddings}   &    \multicolumn{7}{>{\columncolor[gray]{0.9}}c}{BoW}\\
Topics &         \pst &     CTM &    ZSTM &             \pst &    Agg. & Agg.+KM &      KM &  N-PLDA &     LDA &  DBScan \\
\midrule
5  &                &   0.9591 &   0.977* &  \textbf{1.0}* &   0.551 &   0.551 &   0.551 &    0.984 &  0.8398 &        \\
6  &  \textbf{1.0}* &   0.9904 &   0.9776 &           1.0* &  0.4982 &  0.4982 &  0.4982 &   0.9715 &   0.908 &        \\
7  &           1.0* &   0.9583 &   0.9908 &                &  0.4886 &  0.4886 &  0.4886 &  0.9823* &  0.8773 &        \\
10 &                &  0.9728* &   0.9592 &                &  0.6516 &  0.6526 &  0.6569 &  0.9656* &  0.9244 &        \\
11 &                &   0.9646 &  0.9773* &           1.0* &  0.7111 &  0.7105 &  0.7101 &   0.9806 &  0.9313 &        \\
12 &           1.0* &   0.9759 &   0.9644 &                &  0.7579 &  0.7574 &  0.6799 &  0.9829* &  0.9232 &        \\
13 &           1.0* &   0.9387 &    0.974 &           1.0* &  0.7947 &  0.7943 &  0.7905 &   0.9758 &  0.9255 &        \\
15 &                &   0.9641 &  0.9909* &                &   0.743 &  0.7427 &  0.7436 &  0.9819* &  0.9372 &        \\
20 &        0.9994* &   0.9835 &   0.9757 &                &  0.8455 &  0.8448 &  0.8452 &  0.9874* &  0.9415 &        \\
22 &        0.9996* &   0.9844 &   0.9881 &         0.994* &  0.8712 &  0.8705 &  0.8525 &    0.981 &  0.9404 &        \\
25 &                &  0.9872* &   0.9867 &        0.9969* &  0.8819 &  0.8813 &  0.8783 &   0.9781 &  0.9488 &        \\
\bottomrule
\end{tabular}
\caption{IRBO values for Google News data set.}
\label{tab:gn_rbo}

\begin{tabular}{>{\columncolor[gray]{0.9}}l|>{\columncolor[gray]{0.85}}l>{\columncolor[gray]{0.9}}l>{\columncolor[gray]{0.9}}l|>{\columncolor[gray]{0.85}}l>{\columncolor[gray]{0.9}}l>{\columncolor[gray]{0.9}}l>{\columncolor[gray]{0.9}}l>{\columncolor[gray]{0.9}}l>{\columncolor[gray]{0.9}}l>{\columncolor[gray]{0.9}}l}
\toprule
\underline{Input}    &   \multicolumn{3}{>{\columncolor[gray]{0.9}}c|}{SBERT embeddings}   &    \multicolumn{7}{>{\columncolor[gray]{0.9}}c}{BoW}\\
Topics &         \pst &     CTM &    ZSTM &             \pst &    Agg. & Agg.+KM &      KM &  N-PLDA &     LDA &  DBScan \\
\midrule
5  &  \textbf{1.0}* &     1.0* &      1.0 &   0.942 &  0.9371 &  0.9429 &  0.9309 &  \textbf{1.0}* &  0.9511 &         \\
8  &           1.0* &   0.9986 &      1.0 &  0.9897 &  0.9523 &  0.9532 &  0.9359 &           1.0* &  0.9503 &         \\
10 &                &     1.0* &   0.9987 &         &  0.9535 &  0.9543 &  0.9431 &        0.9991* &  0.9469 &         \\
13 &           1.0* &   0.9985 &   0.9987 &   0.995 &  0.9481 &  0.9412 &  0.9395 &        0.9994* &  0.9594 &         \\
15 &           1.0* &   0.9992 &   0.9985 &         &   0.949 &  0.9392 &  0.9482 &        0.9995* &  0.9618 &         \\
18 &                &  0.9991* &   0.9975 &         &  0.9436 &  0.9424 &  0.9516 &         0.999* &   0.968 &  0.9293 \\
20 &                &  0.9984* &   0.9977 &         &  0.9503 &  0.9458 &  0.9277 &        0.9974* &  0.9722 &         \\
23 &           1.0* &   0.9968 &   0.9981 &  0.9961 &  0.9537 &   0.948 &  0.9412 &        0.9982* &  0.9725 &         \\
24 &                &   0.9967 &  0.9983* &  0.9958 &  0.9555 &  0.9499 &  0.9492 &        0.9978* &   0.976 &         \\
25 &           1.0* &    0.997 &   0.9979 &         &  0.9543 &  0.9493 &  0.9482 &        0.9979* &  0.9801 &         \\
\bottomrule
\end{tabular}
\caption{IRBO values for Wiki20K data set.}
\label{tab:db_rbo}

\begin{tabular}{>{\columncolor[gray]{0.9}}l|>{\columncolor[gray]{0.85}}l>{\columncolor[gray]{0.9}}l>{\columncolor[gray]{0.9}}l|>{\columncolor[gray]{0.85}}l>{\columncolor[gray]{0.9}}l>{\columncolor[gray]{0.9}}l>{\columncolor[gray]{0.9}}l>{\columncolor[gray]{0.9}}l>{\columncolor[gray]{0.9}}l>{\columncolor[gray]{0.9}}l}
\toprule
\underline{Input}    &   \multicolumn{3}{>{\columncolor[gray]{0.9}}c|}{SBERT embeddings}   &    \multicolumn{7}{>{\columncolor[gray]{0.9}}c}{BoW}\\
Topics &         \pst &     CTM &    ZSTM &             \pst &    Agg. & Agg.+KM &      KM &  N-PLDA &     LDA &  DBScan \\
\midrule
5  &          &  \textbf{1.0}* &      1.0 &  \textbf{1.0}* &  0.9796 &  0.9796 &  0.9943 &     1.0* &  0.9176 &         \\
6  &          &           1.0* &      1.0 &           1.0* &  0.9864 &  0.9835 &  0.9838 &      1.0 &  0.9283 &         \\
10 &   0.9943 &           1.0* &      1.0 &                &  0.9845 &  0.9819 &  0.9808 &     1.0* &  0.9332 &         \\
11 &          &           1.0* &      1.0 &         0.9985 &  0.9866 &  0.9844 &  0.9814 &  0.9993* &  0.9575 &         \\
15 &          &        0.9992* &   0.9985 &                &  0.9912 &  0.9893 &  0.9756 &     1.0* &  0.9516 &         \\
19 &  0.9988* &         0.9939 &   0.9962 &                &  0.9896 &  0.9899 &  0.9828 &  0.9989* &  0.9671 &         \\
20 &          &         0.9975 &  0.9977* &                &  0.9906 &  0.9909 &  0.9836 &  0.9988* &  0.9687 &         \\
23 &          &         0.9957 &  0.9975* &                &  0.9926 &  0.9925 &   0.986 &  0.9963* &  0.9724 &  0.9075 \\
25 &          &         0.9912 &  0.9955* &                &  0.9932 &  0.9929 &  0.9799 &  0.9977* &  0.9718 &         \\
\bottomrule
\end{tabular}
\caption{IRBO for Tweet data set.}
\label{tab:tw_rbo}
\end{table*}

\begin{table*}[htpb]
\subsection{WECO values}
\label{app_sec_weco}


\centering
\tiny
\begin{tabular}{>{\columncolor[gray]{0.9}}l|>{\columncolor[gray]{0.85}}l>{\columncolor[gray]{0.9}}l>{\columncolor[gray]{0.9}}l|>{\columncolor[gray]{0.85}}l>{\columncolor[gray]{0.9}}l>{\columncolor[gray]{0.9}}l>{\columncolor[gray]{0.9}}l>{\columncolor[gray]{0.9}}l>{\columncolor[gray]{0.9}}l>{\columncolor[gray]{0.9}}l}
\toprule
\underline{Input}    &   \multicolumn{3}{>{\columncolor[gray]{0.9}}c|}{SBERT embeddings}   &    \multicolumn{7}{>{\columncolor[gray]{0.9}}c}{BoW}\\
Topics &         \pst &     CTM &    ZSTM &             \pst &    Agg. & Agg.+KM &      KM &  N-PLDA &     LDA &  DBScan \\
\midrule
5  &  \textbf{0.2174}* &   0.1701 &    0.163 &  \textbf{0.2217}* &    0.183 &  0.1661 &   0.1612 &   0.1701 &  0.1801 &         \\
6  &                   &    0.153 &  0.1888* &            0.1824 &  0.1861* &  0.1821 &   0.1842 &   0.1667 &   0.183 &         \\
7  &           0.2122* &     0.17 &   0.1757 &                   &  0.1805* &  0.1793 &   0.1793 &    0.175 &  0.1763 &         \\
8  &                   &  0.1895* &   0.1873 &            0.1823 &  0.1876* &  0.1831 &   0.1699 &    0.181 &    0.17 &         \\
9  &            0.1963 &  0.2017* &   0.1804 &           0.1931* &    0.189 &  0.1812 &   0.1801 &   0.1848 &  0.1721 &         \\
10 &                   &   0.1845 &  0.2045* &                   &  0.1844* &  0.1736 &     0.17 &   0.1766 &  0.1685 &         \\
\bottomrule
\end{tabular}
\caption{WECO values for Reuters data set.}
\label{tab:re_weco}

\begin{tabular}{>{\columncolor[gray]{0.9}}l|>{\columncolor[gray]{0.85}}l>{\columncolor[gray]{0.9}}l>{\columncolor[gray]{0.9}}l|>{\columncolor[gray]{0.85}}l>{\columncolor[gray]{0.9}}l>{\columncolor[gray]{0.9}}l>{\columncolor[gray]{0.9}}l>{\columncolor[gray]{0.9}}l>{\columncolor[gray]{0.9}}l>{\columncolor[gray]{0.9}}l}
\toprule
\underline{Input}    &   \multicolumn{3}{>{\columncolor[gray]{0.9}}c|}{SBERT embeddings}   &    \multicolumn{7}{>{\columncolor[gray]{0.9}}c}{BoW}\\
Topics &         \pst &     CTM &    ZSTM &             \pst &    Agg. & Agg.+KM &      KM &  N-PLDA &     LDA &  DBScan \\
\midrule
5  &          &   0.2257 &           0.2582* &  0.2527 &  \textbf{0.2622}* &   0.2622 &   0.2622 &  0.2616 &  0.2239 &        \\
6  &   0.2145 &     0.22 &  \textbf{0.2612}* &  0.2273 &           0.2611* &   0.2611 &   0.2611 &  0.2366 &   0.216 &        \\
7  &  0.2162* &   0.1919 &            0.2149 &         &           0.2504* &   0.2504 &   0.2504 &  0.2235 &  0.1978 &        \\
10 &          &   0.2169 &           0.2329* &         &           0.2338* &   0.2338 &   0.2309 &  0.2164 &  0.1898 &        \\
11 &          &  0.2104* &            0.2059 &  0.1966 &            0.2244 &   0.2259 &  0.2268* &  0.1927 &  0.1996 &        \\
12 &   0.2093 &   0.2015 &           0.2179* &         &            0.2373 &  0.2386* &   0.2345 &  0.2311 &  0.2043 &        \\
\bottomrule
\end{tabular}
\caption{WECO values for Google News data set.}
\label{tab:gn_weco}

\begin{tabular}{>{\columncolor[gray]{0.9}}l|>{\columncolor[gray]{0.85}}l>{\columncolor[gray]{0.9}}l>{\columncolor[gray]{0.9}}l|>{\columncolor[gray]{0.85}}l>{\columncolor[gray]{0.9}}l>{\columncolor[gray]{0.9}}l>{\columncolor[gray]{0.9}}l>{\columncolor[gray]{0.9}}l>{\columncolor[gray]{0.9}}l>{\columncolor[gray]{0.9}}l}
\toprule
\underline{Input}    &   \multicolumn{3}{>{\columncolor[gray]{0.9}}c|}{SBERT embeddings}   &    \multicolumn{7}{>{\columncolor[gray]{0.9}}c}{BoW}\\
Topics &         \pst &     CTM &    ZSTM &             \pst &    Agg. & Agg.+KM &      KM &  N-PLDA &     LDA &  DBScan \\
\midrule
5  &                   &  0.2015 &  0.2125* &  0.1864 &    0.257 &  \textbf{0.2608}* &   0.2258 &   0.2239 &  0.1946 &         \\
8  &            0.2138 &  0.1952 &  0.2184* &  0.1748 &   0.2246 &           0.2315* &   0.2249 &   0.2019 &  0.1673 &         \\
10 &                   &  0.2081 &  0.2155* &         &  0.2184* &            0.2101 &   0.2039 &   0.2105 &  0.1701 &         \\
13 &  \textbf{0.2474}* &  0.2172 &   0.2074 &  0.1695 &  0.2135* &            0.2122 &    0.208 &   0.1999 &  0.1706 &         \\
15 &            0.2111 &  0.2071 &  0.2204* &         &   0.2074 &            0.2006 &    0.207 &  0.2135* &  0.1745 &         \\
18 &                   &   0.208 &  0.2232* &         &   0.2027 &            0.1983 &   0.1929 &  0.2239* &  0.1612 &  0.1234 \\
20 &                   &  0.1938 &  0.2095* &         &   0.1994 &           0.2032* &   0.1967 &   0.2032 &  0.1701 &         \\
23 &            0.2054 &  0.2136 &  0.2149* &  0.1766 &   0.2056 &            0.2077 &  0.2126* &   0.2033 &   0.164 &         \\
24 &                   &  0.1981 &  0.2192* &  0.1709 &   0.2026 &            0.2042 &   0.1922 &  0.2053* &  0.1737 &         \\
25 &            0.1993 &  0.1965 &  0.2186* &         &   0.2053 &           0.2091* &   0.1979 &   0.2074 &  0.1668 &         \\
\bottomrule
\end{tabular}
\caption{WECO values for Wiki20K data set.}
\label{tab:db_weco}

\begin{tabular}{>{\columncolor[gray]{0.9}}l|>{\columncolor[gray]{0.85}}l>{\columncolor[gray]{0.9}}l>{\columncolor[gray]{0.9}}l|>{\columncolor[gray]{0.85}}l>{\columncolor[gray]{0.9}}l>{\columncolor[gray]{0.9}}l>{\columncolor[gray]{0.9}}l>{\columncolor[gray]{0.9}}l>{\columncolor[gray]{0.9}}l>{\columncolor[gray]{0.9}}l}
\toprule
\underline{Input}    &   \multicolumn{3}{>{\columncolor[gray]{0.9}}c|}{SBERT embeddings}   &    \multicolumn{7}{>{\columncolor[gray]{0.9}}c}{BoW}\\
Topics &         \pst &     CTM &    ZSTM &             \pst &    Agg. & Agg.+KM &      KM &  N-PLDA &     LDA &  DBScan \\
\midrule
5  &                   &  0.1287* &   0.1226 &         &  0.1445 &           0.1471* &   0.143 &  0.1242 &  0.1066 &         \\
6  &                   &  0.1498* &   0.1175 &  0.1315 &  0.1434 &           0.1499* &  0.1498 &  0.1267 &  0.1124 &         \\
10 &  \textbf{0.1619}* &   0.1475 &   0.1458 &         &  0.1473 &           0.1574* &  0.1562 &  0.1388 &  0.1168 &         \\
11 &                   &  0.1612* &    0.147 &  0.1278 &  0.1619 &  \textbf{0.1712}* &  0.1348 &  0.1374 &   0.121 &         \\
15 &                   &  0.1599* &    0.156 &         &  0.1531 &           0.1602* &  0.1441 &   0.143 &  0.1109 &         \\
19 &             0.138 &   0.1521 &  0.1522* &         &  0.1515 &           0.1572* &  0.1533 &  0.1405 &  0.1049 &         \\
20 &                   &   0.1487 &  0.1577* &         &  0.1523 &           0.1573* &  0.1553 &  0.1386 &   0.114 &         \\
23 &                   &   0.1563 &  0.1617* &         &  0.1578 &           0.1623* &  0.1574 &  0.1509 &  0.1208 &  0.1034 \\
25 &                   &   0.1472 &  0.1511* &         &   0.157 &           0.1614* &  0.1553 &  0.1452 &  0.1121 &         \\
\bottomrule
\end{tabular}
\caption{WECO values for Tweet data set.}
\label{tab:tw_weco}
\end{table*}

\begin{table*}[htpb]
\subsection{Specificity values}
\label{app_sec_spec}


\centering
\tiny
\begin{tabular}{>{\columncolor[gray]{0.9}}l|>{\columncolor[gray]{0.85}}l>{\columncolor[gray]{0.9}}l>{\columncolor[gray]{0.9}}l|>{\columncolor[gray]{0.85}}l>{\columncolor[gray]{0.9}}l>{\columncolor[gray]{0.9}}l>{\columncolor[gray]{0.9}}l>{\columncolor[gray]{0.9}}l>{\columncolor[gray]{0.9}}l>{\columncolor[gray]{0.9}}l}
\toprule
\underline{Input}    &   \multicolumn{3}{>{\columncolor[gray]{0.9}}c|}{SBERT embeddings}   &    \multicolumn{7}{>{\columncolor[gray]{0.9}}c}{BoW}\\
Topics &         \pst &     CTM &    ZSTM &             \pst &    Agg. & Agg.+KM &      KM &  N-PLDA &     LDA &  DBScan \\
\midrule
5  &            1.4708 &  1.6094* &   1.6094 &    1.443 &  0.8017 &   0.873 &  0.8835 &           1.6094* &  1.0062 &         \\
6  &                   &   1.7224 &  1.7686* &   1.7224 &  0.9095 &   0.914 &   0.767 &           1.7917* &  1.0947 &         \\
7  &            1.7874 &   1.8666 &  1.8864* &          &  1.0166 &  0.9065 &  0.8788 &           1.9458* &  1.2851 &         \\
8  &                   &   1.9581 &  2.0447* &  2.0794* &  1.0179 &   0.904 &  0.9491 &            2.0274 &    1.35 &         \\
9  &           2.1201* &   2.0123 &   2.0585 &  2.1971* &  1.1531 &  1.0179 &  0.9222 &            2.1047 &  1.5212 &         \\
10 &                   &   2.1084 &  2.1777* &          &  1.2001 &  1.0347 &  1.0897 &           2.1223* &  1.5763 &         \\
\bottomrule
\end{tabular}
\caption{Specificity values for Reuters data set.}
\label{tab:re_specif}

\begin{tabular}{>{\columncolor[gray]{0.9}}l|>{\columncolor[gray]{0.85}}l>{\columncolor[gray]{0.9}}l>{\columncolor[gray]{0.9}}l|>{\columncolor[gray]{0.85}}l>{\columncolor[gray]{0.9}}l>{\columncolor[gray]{0.9}}l>{\columncolor[gray]{0.9}}l>{\columncolor[gray]{0.9}}l>{\columncolor[gray]{0.9}}l>{\columncolor[gray]{0.9}}l}
\toprule
\underline{Input}    &   \multicolumn{3}{>{\columncolor[gray]{0.9}}c|}{SBERT embeddings}   &    \multicolumn{7}{>{\columncolor[gray]{0.9}}c}{BoW}\\
Topics &         \pst &     CTM &    ZSTM &             \pst &    Agg. & Agg.+KM &      KM &  N-PLDA &     LDA &  DBScan \\
\midrule
5  &                   &   1.4326 &  1.4985* &           1.6094* &  0.8331 &  0.8331 &  0.8331 &   1.5262 &  1.2071 &        \\
6  &           1.7917* &   1.7224 &   1.6531 &           1.7917* &  0.8726 &  0.8726 &  0.8726 &     1.63 &  1.4739 &        \\
7  &           1.9458* &   1.7676 &   1.8666 &                   &  0.9169 &  0.9169 &  0.9169 &  1.8468* &  1.5075 &        \\
10 &                   &  2.1118* &   2.0043 &                   &  1.2688 &  1.2827 &  1.3104 &  2.0702* &  1.9318 &        \\
11 &                   &   2.1567 &  2.2228* &           2.3978* &  1.4329 &  1.4203 &  1.4203 &   2.2922 &  2.0403 &        \\
12 &           2.4848* &   2.3187 &   2.2045 &                   &  1.5772 &  1.5656 &  1.3979 &  2.3534* &  2.0444 &        \\
\bottomrule
\end{tabular}
\caption{Specificity values for Google News data set.}
\label{tab:gn_specif}

\begin{tabular}{>{\columncolor[gray]{0.9}}l|>{\columncolor[gray]{0.85}}l>{\columncolor[gray]{0.9}}l>{\columncolor[gray]{0.9}}l|>{\columncolor[gray]{0.85}}l>{\columncolor[gray]{0.9}}l>{\columncolor[gray]{0.9}}l>{\columncolor[gray]{0.9}}l>{\columncolor[gray]{0.9}}l>{\columncolor[gray]{0.9}}l>{\columncolor[gray]{0.9}}l}
\toprule
\underline{Input}    &   \multicolumn{3}{>{\columncolor[gray]{0.9}}c|}{SBERT embeddings}   &    \multicolumn{7}{>{\columncolor[gray]{0.9}}c}{BoW}\\
Topics &         \pst &     CTM &    ZSTM &             \pst &    Agg. & Agg.+KM &      KM &  N-PLDA &     LDA &  DBScan \\
\midrule
5  &                   &  1.6094* &   1.6094 &   1.443 &  1.3321 &  1.3599 &  1.2662 &           1.6094* &  1.4049 &         \\
8  &           2.0794* &    2.062 &   2.0794 &    2.01 &  1.7544 &  1.7652 &  1.6547 &           2.0794* &  1.7685 &         \\
10 &                   &  2.3025* &   2.2886 &         &   1.935 &  1.9325 &  1.8388 &           2.2886* &  1.9507 &         \\
13 &           2.5648* &   2.5435 &   2.5435 &  2.5008 &  2.0675 &  2.0047 &  2.0144 &           2.5542* &  2.1643 &         \\
15 &           2.7079* &   2.6894 &   2.6894 &         &  2.1729 &  2.0862 &  2.2128 &           2.6987* &   2.293 &         \\
18 &                   &  2.8748* &   2.8671 &         &  2.2281 &  2.2029 &  2.3116 &           2.8748* &  2.4555 &  2.0711 \\
20 &                   &   2.9539 &  2.9609* &         &  2.3723 &  2.3081 &  2.1905 &            2.947* &  2.6156 &         \\
23 &           3.1353* &   3.0486 &    3.087 &  3.0538 &  2.4895 &  2.4021 &  2.3161 &           3.0931* &  2.7141 &         \\
24 &                   &    3.097 &  3.1201* &  3.1006 &  2.5302 &  2.4398 &  2.4627 &           3.1258* &  2.7608 &         \\
25 &  \textbf{3.2186}* &   3.1389 &   3.1632 &         &  2.5498 &  2.4629 &  2.4282 &  \textbf{3.1666}* &   2.845 &         \\
\bottomrule
\end{tabular}
\caption{Specificity values for Wiki20K data set.}
\label{tab:db_specif}

\begin{tabular}{>{\columncolor[gray]{0.9}}l|>{\columncolor[gray]{0.85}}l>{\columncolor[gray]{0.9}}l>{\columncolor[gray]{0.9}}l|>{\columncolor[gray]{0.85}}l>{\columncolor[gray]{0.9}}l>{\columncolor[gray]{0.9}}l>{\columncolor[gray]{0.9}}l>{\columncolor[gray]{0.9}}l>{\columncolor[gray]{0.9}}l>{\columncolor[gray]{0.9}}l}
\toprule
\underline{Input}    &   \multicolumn{3}{>{\columncolor[gray]{0.9}}c|}{SBERT embeddings}   &    \multicolumn{7}{>{\columncolor[gray]{0.9}}c}{BoW}\\
Topics &         \pst &     CTM &    ZSTM &             \pst &    Agg. & Agg.+KM &      KM &  N-PLDA &     LDA &  DBScan \\
\midrule
5  &          &  1.6094* &            1.6094 &          &  1.5262 &  1.5262 &  1.5817 &           1.6094* &  1.3494 &         \\
6  &          &  1.7917* &            1.7917 &  1.7999* &  1.7224 &  1.6993 &  1.6993 &            1.7917 &  1.4826 &         \\
10 &   2.2332 &  2.3025* &            2.3025 &          &   2.162 &  2.1257 &  2.1777 &           2.3025* &    1.85 &         \\
11 &          &  2.3978* &            2.3978 &   2.3726 &  2.2575 &  2.2244 &  2.2401 &           2.3852* &  2.1087 &         \\
15 &          &  2.6894* &            2.6802 &          &    2.59 &  2.5566 &  2.4594 &           2.7079* &  2.3294 &         \\
19 &  2.9304* &   2.8612 &            2.8786 &          &  2.7828 &   2.781 &  2.6879 &           2.9224* &   2.583 &         \\
20 &          &  2.9539* &             2.947 &          &  2.8421 &  2.8404 &  2.7451 &           2.9747* &  2.6354 &         \\
23 &          &   3.0569 &            3.069* &          &  2.9898 &  2.9763 &  2.8949 &           3.0569* &   2.772 &  1.9899 \\
25 &          &   3.0682 &  \textbf{3.1355}* &          &  3.0703 &  3.0523 &  2.8883 &  \textbf{3.1632}* &   2.818 &         \\
\bottomrule
\end{tabular}
\caption{Specificity values for Tweet data set.}
\label{tab:tw_specif}
\end{table*}

\begin{table*}[htpb]
\subsection{Dissimilarity values}
\label{app_sec_diss}


\centering
\tiny
\begin{tabular}{>{\columncolor[gray]{0.9}}l|>{\columncolor[gray]{0.85}}l>{\columncolor[gray]{0.9}}l>{\columncolor[gray]{0.9}}l|>{\columncolor[gray]{0.85}}l>{\columncolor[gray]{0.9}}l>{\columncolor[gray]{0.9}}l>{\columncolor[gray]{0.9}}l>{\columncolor[gray]{0.9}}l>{\columncolor[gray]{0.9}}l>{\columncolor[gray]{0.9}}l}
\toprule
\underline{Input}    &   \multicolumn{3}{>{\columncolor[gray]{0.9}}c|}{SBERT embeddings}   &    \multicolumn{7}{>{\columncolor[gray]{0.9}}c}{BoW}\\
Topics &         \pst &     CTM &    ZSTM &             \pst &    Agg. & Agg.+KM &      KM &  N-PLDA &     LDA &  DBScan \\
\midrule
5  &           0.3657* &   0.3536 &   0.3536 &           0.3657* &   0.364 &   0.3317 &   0.3202 &   0.3536 &   0.3518 &        \\
6  &                   &  0.3499* &   0.3476 &           0.3533* &  0.3262 &   0.3187 &   0.3237 &   0.3464 &   0.3441 &        \\
7  &           0.3512* &   0.3448 &    0.344 &                   &  0.3215 &   0.3153 &   0.3091 &  0.3416* &   0.3391 &        \\
8  &                   &  0.3423* &   0.3393 &            0.3381 &  0.3307 &   0.3175 &   0.3117 &  0.3399* &   0.3387 &        \\
9  &            0.3387 &   0.341* &   0.3396 &            0.3354 &  0.3298 &   0.3062 &   0.3092 &  0.3382* &   0.3345 &        \\
10 &                   &  0.3385* &   0.3367 &                   &  0.3304 &    0.301 &   0.3243 &  0.3381* &   0.3239 &        \\
\bottomrule
\end{tabular}
\caption{Dissimilarity values for Reuters data set.}
\label{tab:re_dissim}

\begin{tabular}{>{\columncolor[gray]{0.9}}l|>{\columncolor[gray]{0.85}}l>{\columncolor[gray]{0.9}}l>{\columncolor[gray]{0.9}}l|>{\columncolor[gray]{0.85}}l>{\columncolor[gray]{0.9}}l>{\columncolor[gray]{0.9}}l>{\columncolor[gray]{0.9}}l>{\columncolor[gray]{0.9}}l>{\columncolor[gray]{0.9}}l>{\columncolor[gray]{0.9}}l}
\toprule
\underline{Input}    &   \multicolumn{3}{>{\columncolor[gray]{0.9}}c|}{SBERT embeddings}   &    \multicolumn{7}{>{\columncolor[gray]{0.9}}c}{BoW}\\
Topics &         \pst &     CTM &    ZSTM &             \pst &    Agg. & Agg.+KM &      KM &  N-PLDA &     LDA &  DBScan \\
\midrule
5  &                   &  0.3657* &   0.3606 &            0.3588 &  0.4153* &   0.4153 &   0.4153 &  0.3588 &  0.3824 &        \\
6  &            0.3499 &   0.3499 &  0.3533* &            0.3464 &  0.4118* &   0.4118 &   0.4118 &  0.3544 &  0.3655 &        \\
7  &            0.3432 &  0.3488* &   0.3448 &                   &  0.4062* &   0.4062 &   0.4062 &  0.3456 &  0.3651 &        \\
10 &                   &   0.3388 &  0.3435* &                   &  0.3794* &   0.3791 &   0.3784 &  0.3399 &  0.3475 &        \\
11 &                   &  0.3379* &   0.3367 &            0.3329 &   0.3701 &  0.3704* &   0.3704 &  0.3344 &  0.3441 &        \\
12 &            0.3303 &    0.335 &  0.3377* &                   &   0.3602 &   0.3604 &  0.3693* &  0.3333 &  0.3447 &        \\
\bottomrule
\end{tabular}
\caption{Dissimilarity values for Google News data set.}
\label{tab:gn_dissim}

\begin{tabular}{>{\columncolor[gray]{0.9}}l|>{\columncolor[gray]{0.85}}l>{\columncolor[gray]{0.9}}l>{\columncolor[gray]{0.9}}l|>{\columncolor[gray]{0.85}}l>{\columncolor[gray]{0.9}}l>{\columncolor[gray]{0.9}}l>{\columncolor[gray]{0.9}}l>{\columncolor[gray]{0.9}}l>{\columncolor[gray]{0.9}}l>{\columncolor[gray]{0.9}}l}
\toprule
\underline{Input}    &   \multicolumn{3}{>{\columncolor[gray]{0.9}}c|}{SBERT embeddings}   &    \multicolumn{7}{>{\columncolor[gray]{0.9}}c}{BoW}\\
Topics &         \pst &     CTM &    ZSTM &             \pst &    Agg. & Agg.+KM &      KM &  N-PLDA &     LDA &  DBScan \\
\midrule
5  &         &  \textbf{0.3536}* &   0.3536 &  \textbf{0.364}* &   0.3446 &   0.3518 &    0.341 &  0.3536 &   0.3588 &         \\
8  &  0.3381 &           0.3387* &   0.3381 &           0.3423 &    0.332 &    0.332 &    0.335 &  0.3381 &  0.3429* &         \\
10 &         &            0.3333 &  0.3337* &                  &    0.341 &   0.3385 &  0.3414* &  0.3337 &   0.3341 &         \\
13 &  0.3291 &           0.3296* &   0.3296 &           0.3389 &   0.3391 &  0.3395* &   0.3377 &  0.3294 &   0.3285 &         \\
15 &  0.3273 &           0.3276* &   0.3276 &                  &  0.3375* &   0.3372 &   0.3341 &  0.3275 &   0.3349 &         \\
18 &         &            0.3256 &  0.3257* &                  &  0.3357* &   0.3346 &   0.3293 &  0.3256 &   0.3261 &  0.3341 \\
20 &         &            0.325* &   0.3249 &                  &    0.333 &   0.3327 &  0.3388* &   0.325 &   0.3292 &         \\
23 &  0.3235 &           0.3243* &   0.3238 &           0.3251 &   0.3315 &    0.332 &  0.3322* &  0.3238 &   0.3298 &         \\
24 &         &           0.3238* &   0.3236 &          0.3493* &   0.3311 &   0.3316 &   0.3299 &  0.3236 &   0.3257 &         \\
25 &  0.3227 &           0.3236* &   0.3233 &                  &   0.3311 &  0.3315* &   0.3254 &  0.3233 &   0.3237 &         \\
\bottomrule
\end{tabular}
\caption{Dissimilarity values for Wiki20K data set.}
\label{tab:db_dissim}

\begin{tabular}{>{\columncolor[gray]{0.9}}l|>{\columncolor[gray]{0.85}}l>{\columncolor[gray]{0.9}}l>{\columncolor[gray]{0.9}}l|>{\columncolor[gray]{0.85}}l>{\columncolor[gray]{0.9}}l>{\columncolor[gray]{0.9}}l>{\columncolor[gray]{0.9}}l>{\columncolor[gray]{0.9}}l>{\columncolor[gray]{0.9}}l>{\columncolor[gray]{0.9}}l}
\toprule
\underline{Input}    &   \multicolumn{3}{>{\columncolor[gray]{0.9}}c|}{SBERT embeddings}   &    \multicolumn{7}{>{\columncolor[gray]{0.9}}c}{BoW}\\
Topics &         \pst &     CTM &    ZSTM &             \pst &    Agg. & Agg.+KM &      KM &  N-PLDA &     LDA &  DBScan \\
\midrule
5  &                   &  0.3536* &   0.3536 &                   &  0.3588* &   0.3588 &   0.3553 &  0.3536 &   0.3536 &          \\
6  &                   &  0.3464* &   0.3464 &  \textbf{0.4036}* &   0.3499 &    0.351 &    0.351 &  0.3464 &   0.3555 &          \\
10 &           0.3352* &   0.3333 &   0.3333 &                   &   0.3377 &  0.3385* &   0.3367 &  0.3333 &   0.3344 &          \\
11 &                   &  0.3317* &   0.3317 &            0.3323 &   0.3356 &   0.3362 &   0.3362 &   0.332 &   0.337* &          \\
15 &                   &   0.3276 &  0.3278* &                   &   0.3298 &   0.3303 &   0.332* &  0.3273 &    0.332 &          \\
19 &  \textbf{0.4105}* &    0.326 &   0.3257 &                   &   0.3274 &   0.3273 &  0.3286* &  0.3252 &   0.3263 &          \\
20 &                   &   0.325* &    0.325 &                   &   0.3267 &   0.3266 &   0.3277 &  0.3247 &  0.3289* &          \\
23 &                   &  0.3242* &    0.324 &                   &   0.3251 &   0.3252 &   0.3263 &  0.3242 &   0.3288 &  0.3468* \\
25 &                   &  0.3243* &   0.3236 &                   &   0.3245 &   0.3246 &   0.3267 &  0.3233 &  0.3289* &          \\
\bottomrule
\end{tabular}
\caption{Dissimilarity values for Tweet data set.}
\label{tab:tw_dissim}
\end{table*}

\begin{table*}[htpb]
\subsection{Comparison with Top2Vec and Sia et al. (2020) method}
\label{app_sec_tired}
\centering
\tiny
\begin{minipage}[l]{0.99\columnwidth}
\begin{tabular}{>{\columncolor[gray]{0.9}}l|>{\columncolor[gray]{0.85}}l>{\columncolor[gray]{0.9}}l|>{\columncolor[gray]{0.9}}l>{\columncolor[gray]{0.9}}l>{\columncolor[gray]{0.9}}l}
\toprule
\underline{Input}    &   \multicolumn{2}{>{\columncolor[gray]{0.9}}c|}{Doc2Vec embeddings}   &    \multicolumn{3}{>{\columncolor[gray]{0.9}}c}{Word2vec embeddings}\\
{} &   \pst &               T2V &               KM & Spherical-KM &      GMM \\
\midrule
5   &           0.7005* &          &    0.356 &            0.3809 &  0.4485* \\
6   &                   &          &   0.4154 &           0.4332* &    0.431 \\
7   &                   &          &   0.4515 &  \textbf{0.4702}* &   0.4492 \\
8   &  \textbf{0.7044}* &          &   0.4043 &           0.4361* &   0.4333 \\
9   &                   &          &    0.405 &           0.4204* &   0.3755 \\
10  &                   &          &  0.4187* &            0.4186 &   0.3843 \\
11  &                   &          &   0.433* &            0.4296 &   0.4102 \\
13  &                   &          &   0.4095 &            0.3794 &  0.4303* \\
15  &           0.6296* &          &   0.4121 &            0.3772 &  0.4622* \\
16  &                   &          &   0.4214 &            0.3966 &  0.4465* \\
17  &                   &          &   0.3809 &            0.3964 &    0.45* \\
20  &                   &          &   0.3931 &            0.3935 &  0.4418* \\
21  &                   &          &   0.3905 &            0.4027 &  0.4353* \\
25  &           0.6744* &          &  0.4063* &            0.3661 &   0.4004 \\
114 &                   &  0.6546* &          &                   &          \\
\bottomrule
\end{tabular}
\caption{Coherence values for Reuters data set.}
\label{tab:re_cv_tired}
\end{minipage}
\qquad
\begin{minipage}[l]{0.99\columnwidth}
\begin{tabular}{>{\columncolor[gray]{0.9}}l|>{\columncolor[gray]{0.85}}l>{\columncolor[gray]{0.9}}l|>{\columncolor[gray]{0.9}}l>{\columncolor[gray]{0.9}}l>{\columncolor[gray]{0.9}}l}
\toprule
\underline{Input}    &   \multicolumn{2}{>{\columncolor[gray]{0.9}}c|}{Doc2Vec embeddings}   &    \multicolumn{3}{>{\columncolor[gray]{0.9}}c}{Word2vec embeddings}\\
{} &   \pst &               T2V &               KM & Spherical-KM &      GMM \\
\midrule
2  &          &  \textbf{0.7799}* &                  &              &          \\
4  &   0.701* &                   &                  &              &          \\
5  &          &                   &          0.4237* &              &    0.391 \\
6  &          &                   &           0.3765 &      0.4836* &   0.4322 \\
7  &   0.711* &                   &           0.5109 &      0.5307* &   0.4022 \\
10 &          &                   &          0.5008* &       0.4967 &   0.4731 \\
11 &          &                   &           0.4748 &       0.491* &   0.4734 \\
12 &          &                   &          0.5215* &       0.5144 &   0.4736 \\
13 &  0.7014* &                   &           0.4626 &      0.5006* &   0.4934 \\
15 &          &                   &           0.5259 &      0.5363* &    0.486 \\
20 &  0.7022* &                   &          0.5654* &       0.5385 &   0.5324 \\
22 &          &                   &           0.5052 &       0.5147 &  0.5297* \\
25 &          &                   &  \textbf{0.568}* &       0.5297 &   0.5286 \\
\bottomrule
\end{tabular}
\caption{Coherence values for Google News data set.}
\label{tab:gn_cv_tired}
\end{minipage}

\vspace{2mm}

\begin{minipage}[l]{0.99\columnwidth}
\begin{tabular}{>{\columncolor[gray]{0.9}}l|>{\columncolor[gray]{0.85}}l>{\columncolor[gray]{0.9}}l|>{\columncolor[gray]{0.9}}l>{\columncolor[gray]{0.9}}l>{\columncolor[gray]{0.9}}l}
\toprule
\underline{Input}    &   \multicolumn{2}{>{\columncolor[gray]{0.9}}c|}{Doc2Vec embeddings}   &    \multicolumn{3}{>{\columncolor[gray]{0.9}}c}{Word2vec embeddings}\\
{} &   \pst &               T2V &               KM & Spherical-KM &      GMM \\
\midrule
5   &  \textbf{0.2013}* &          &  -0.2067 &           -0.0966* &   -0.1316 \\
6   &                   &          &  -0.1502 &            -0.063* &   -0.1563 \\
7   &                   &          &  -0.1527 &  \textbf{-0.0542}* &   -0.1332 \\
8   &           0.1981* &          &  -0.1377 &           -0.0785* &   -0.1056 \\
9   &                   &          &  -0.1395 &           -0.1011* &   -0.1284 \\
10  &                   &          &  -0.1335 &           -0.1145* &   -0.1312 \\
11  &                   &          &  -0.1345 &           -0.0869* &   -0.1274 \\
13  &                   &          &  -0.1225 &            -0.1312 &  -0.0922* \\
15  &           0.1216* &          &  -0.1384 &            -0.1345 &  -0.0925* \\
16  &                   &          &  -0.113* &            -0.1373 &    -0.113 \\
17  &                   &          &  -0.1565 &            -0.1219 &  -0.0966* \\
20  &                   &          &   -0.137 &            -0.1445 &  -0.1016* \\
21  &                   &          &  -0.1331 &             -0.132 &  -0.0874* \\
25  &           0.1777* &          &  -0.124* &             -0.152 &    -0.129 \\
114 &                   &  0.0994* &          &                    &           \\
\bottomrule
\end{tabular}
\caption{NPMI values for Reuters data set.}
\label{tab:re_npmi_tired}
\end{minipage}
\qquad
\begin{minipage}[l]{0.99\columnwidth}
\begin{tabular}{>{\columncolor[gray]{0.9}}l|>{\columncolor[gray]{0.85}}l>{\columncolor[gray]{0.9}}l|>{\columncolor[gray]{0.9}}l>{\columncolor[gray]{0.9}}l>{\columncolor[gray]{0.9}}l}
\toprule
\underline{Input}    &   \multicolumn{2}{>{\columncolor[gray]{0.9}}c|}{Doc2Vec embeddings}   &    \multicolumn{3}{>{\columncolor[gray]{0.9}}c}{Word2vec embeddings}\\
{} &   \pst &               T2V &               KM & Spherical-KM &      GMM \\
\midrule
2  &          &  \textbf{0.1873}* &          &                   &           \\
4  &  0.1466* &                   &          &                   &           \\
5  &          &                   &   -0.153 &                   &  -0.1215* \\
6  &          &                   &    -0.13 &          -0.0747* &   -0.0919 \\
7  &  0.1595* &                   &  -0.0556 &          -0.0294* &   -0.1073 \\
10 &          &                   &  -0.0705 &           -0.0626 &  -0.0424* \\
11 &          &                   &  -0.0619 &           -0.035* &   -0.0442 \\
12 &          &                   &  -0.0497 &           -0.0441 &  -0.0402* \\
13 &  0.1645* &                   &  -0.0519 &           -0.0389 &  -0.0268* \\
15 &          &                   &  -0.0438 &  \textbf{0.0091}* &   -0.0418 \\
20 &  0.1514* &                   &   -0.011 &           -0.0156 &  -0.0087* \\
22 &          &                   &  -0.0671 &           -0.0325 &  -0.0179* \\
25 &          &                   &   -0.018 &          -0.0146* &   -0.0204 \\
\bottomrule
\end{tabular}
\caption{NPMI values for Google News data set.}
\label{tab:gn_npmi_tired}
\end{minipage}

\vspace{2mm}

\begin{minipage}[l]{0.99\columnwidth}
\begin{tabular}{>{\columncolor[gray]{0.9}}l|>{\columncolor[gray]{0.85}}l>{\columncolor[gray]{0.9}}l|>{\columncolor[gray]{0.9}}l>{\columncolor[gray]{0.9}}l>{\columncolor[gray]{0.9}}l}
\toprule
\underline{Input}    &   \multicolumn{2}{>{\columncolor[gray]{0.9}}c|}{Doc2Vec embeddings}   &    \multicolumn{3}{>{\columncolor[gray]{0.9}}c}{Word2vec embeddings}\\
{} &   \pst &               T2V &               KM & Spherical-KM &      GMM \\
\midrule
5   &  \textbf{1.0}* &          &  \textbf{1.0}* &          1.0 &     1.0 \\
6   &                &          &           1.0* &          1.0 &     1.0 \\
7   &                &          &           1.0* &          1.0 &     1.0 \\
8   &        0.9955* &          &           1.0* &          1.0 &     1.0 \\
9   &                &          &           1.0* &          1.0 &     1.0 \\
10  &                &          &           1.0* &          1.0 &     1.0 \\
11  &                &          &           1.0* &          1.0 &     1.0 \\
13  &                &          &           1.0* &          1.0 &     1.0 \\
15  &        0.9936* &          &           1.0* &          1.0 &     1.0 \\
16  &                &          &           1.0* &          1.0 &     1.0 \\
17  &                &          &           1.0* &          1.0 &     1.0 \\
20  &                &          &           1.0* &          1.0 &     1.0 \\
21  &                &          &           1.0* &          1.0 &  0.9998 \\
25  &        0.9957* &          &           1.0* &          1.0 &  0.9999 \\
114 &                &  0.9811* &                &              &         \\
\bottomrule
\end{tabular}
\caption{IRBO values for Reuters data set.}
\label{tab:re_rbo_tired}
\end{minipage}
\qquad
\begin{minipage}[l]{0.99\columnwidth}
\begin{tabular}{>{\columncolor[gray]{0.9}}l|>{\columncolor[gray]{0.85}}l>{\columncolor[gray]{0.9}}l|>{\columncolor[gray]{0.9}}l>{\columncolor[gray]{0.9}}l>{\columncolor[gray]{0.9}}l}
\toprule
\underline{Input}    &   \multicolumn{2}{>{\columncolor[gray]{0.9}}c|}{Doc2Vec embeddings}   &    \multicolumn{3}{>{\columncolor[gray]{0.9}}c}{Word2vec embeddings}\\
{} &   \pst &               T2V &               KM & Spherical-KM &      GMM \\
\midrule
2  &  \textbf{1.0}* &  1.0* &                &              &         \\
4  &           1.0* &       &                &              &         \\
5  &                &       &  \textbf{1.0}* &              &     1.0 \\
6  &                &       &           1.0* &          1.0 &     1.0 \\
7  &           1.0* &       &           1.0* &          1.0 &     1.0 \\
10 &                &       &           1.0* &          1.0 &     1.0 \\
11 &                &       &           1.0* &          1.0 &     1.0 \\
12 &                &       &           1.0* &          1.0 &     1.0 \\
13 &           1.0* &       &           1.0* &          1.0 &     1.0 \\
15 &                &       &           1.0* &          1.0 &     1.0 \\
20 &           1.0* &       &           1.0* &          1.0 &     1.0 \\
22 &                &       &           1.0* &          1.0 &  0.9991 \\
25 &                &       &           1.0* &          1.0 &  0.9983 \\
\bottomrule
\end{tabular}
\caption{IRBO values for Google News data set.}
\label{tab:gn_rbo_tired}
\end{minipage}

\vspace{2mm}

\begin{minipage}[l]{0.99\columnwidth}
\begin{tabular}{>{\columncolor[gray]{0.9}}l|>{\columncolor[gray]{0.85}}l>{\columncolor[gray]{0.9}}l|>{\columncolor[gray]{0.9}}l>{\columncolor[gray]{0.9}}l>{\columncolor[gray]{0.9}}l}
\toprule
\underline{Input}    &   \multicolumn{2}{>{\columncolor[gray]{0.9}}c|}{Doc2Vec embeddings}   &    \multicolumn{3}{>{\columncolor[gray]{0.9}}c}{Word2vec embeddings}\\
{} &   \pst &               T2V &               KM & Spherical-KM &      GMM \\
\midrule
5   &           0.1666* &         &  0.2594 &           0.3847* &   0.3187 \\
6   &                   &         &  0.2951 &            0.405* &   0.3168 \\
7   &                   &         &   0.264 &           0.3975* &   0.3514 \\
8   &  \textbf{0.2023}* &         &  0.3026 &  \textbf{0.4196}* &    0.332 \\
9   &                   &         &  0.2755 &           0.3949* &   0.3185 \\
10  &                   &         &  0.2815 &           0.3976* &   0.3333 \\
11  &                   &         &  0.2951 &           0.3945* &   0.3287 \\
13  &                   &         &  0.2764 &           0.3892* &   0.3637 \\
15  &           0.1735* &         &  0.2844 &            0.3772 &  0.3902* \\
16  &                   &         &  0.3211 &           0.3903* &   0.3835 \\
17  &                   &         &  0.3017 &            0.3769 &   0.379* \\
20  &                   &         &  0.3005 &            0.3689 &  0.3932* \\
21  &                   &         &  0.2927 &            0.3811 &  0.3907* \\
25  &           0.1441* &         &  0.3263 &            0.3641 &  0.3877* \\
114 &                   &  0.164* &         &                   &          \\
\bottomrule
\end{tabular}
\caption{WECO values for Reuters data set.}
\label{tab:re_weco_tired}
\end{minipage}
\qquad
\begin{minipage}[l]{0.99\columnwidth}
\begin{tabular}{>{\columncolor[gray]{0.9}}l|>{\columncolor[gray]{0.85}}l>{\columncolor[gray]{0.9}}l|>{\columncolor[gray]{0.9}}l>{\columncolor[gray]{0.9}}l>{\columncolor[gray]{0.9}}l}
\toprule
\underline{Input}    &   \multicolumn{2}{>{\columncolor[gray]{0.9}}c|}{Doc2Vec embeddings}   &    \multicolumn{3}{>{\columncolor[gray]{0.9}}c}{Word2vec embeddings}\\
{} &   \pst &               T2V &               KM & Spherical-KM &      GMM \\
\midrule
2  &          &  \textbf{0.3007}* &         &                   &          \\
4  &  0.2259* &                   &         &                   &          \\
5  &          &                   &   0.237 &                   &  0.2923* \\
6  &          &                   &  0.2399 &  \textbf{0.3808}* &   0.3096 \\
7  &  0.2302* &                   &  0.2868 &           0.3622* &   0.2968 \\
10 &          &                   &  0.2722 &            0.3525 &  0.3659* \\
11 &          &                   &  0.2955 &            0.3418 &  0.3539* \\
12 &          &                   &   0.283 &            0.3566 &  0.3653* \\
13 &  0.2412* &                   &  0.3054 &            0.3538 &  0.3697* \\
15 &          &                   &  0.2863 &           0.3632* &   0.3478 \\
20 &  0.2054* &                   &  0.3276 &           0.3492* &    0.341 \\
22 &          &                   &  0.2773 &            0.3447 &  0.3507* \\
25 &          &                   &  0.3139 &            0.3198 &  0.3556* \\
\bottomrule
\end{tabular}
\caption{WECO values for Google News data set.}
\label{tab:gn_weco_tired}
\end{minipage}
\end{table*}

\begin{table*}
\centering
\tiny
\begin{minipage}[l]{0.99\columnwidth}
\begin{tabular}{>{\columncolor[gray]{0.9}}l|>{\columncolor[gray]{0.85}}l>{\columncolor[gray]{0.9}}l|>{\columncolor[gray]{0.9}}l>{\columncolor[gray]{0.9}}l>{\columncolor[gray]{0.9}}l}
\toprule
\underline{Input}    &   \multicolumn{2}{>{\columncolor[gray]{0.9}}c|}{Doc2Vec embeddings}   &    \multicolumn{3}{>{\columncolor[gray]{0.9}}c}{Word2vec embeddings}\\
{} &   \pst &               T2V &               KM & Spherical-KM &      GMM \\
\midrule
5   &           1.6094* &         &           1.6094* &       1.6094 &   1.6094 \\
6   &                   &         &           1.7917* &       1.7917 &   1.7917 \\
7   &                   &         &           1.9458* &       1.9458 &   1.9458 \\
8   &           2.0436* &         &           2.0794* &       2.0794 &   2.0794 \\
9   &                   &         &           2.1971* &       2.1971 &   2.1971 \\
10  &                   &         &           2.3025* &       2.3025 &   2.3025 \\
11  &                   &         &           2.3978* &       2.3978 &   2.3978 \\
13  &                   &         &            2.4848 &      2.5648* &   2.5648 \\
15  &           2.6071* &         &            2.6389 &      2.7079* &   2.6389 \\
16  &                   &         &           2.7724* &       2.7724 &   2.7079 \\
17  &                   &         &           2.7724* &       2.7079 &   2.7724 \\
20  &                   &         &            2.9442 &       2.9442 &  2.9955* \\
21  &                   &         &           3.0443* &       2.9955 &   3.0377 \\
25  &  \textbf{3.1217}* &         &  \textbf{3.2308}* &       3.1778 &    3.172 \\
114 &                   &  3.091* &                   &              &          \\
\bottomrule
\end{tabular}
\caption{Specificity values for Reuters data set.}
\label{tab:re_specif_tired}
\end{minipage}
\qquad
\begin{minipage}[l]{0.99\columnwidth}
\begin{tabular}{>{\columncolor[gray]{0.9}}l|>{\columncolor[gray]{0.85}}l>{\columncolor[gray]{0.9}}l|>{\columncolor[gray]{0.9}}l>{\columncolor[gray]{0.9}}l>{\columncolor[gray]{0.9}}l}
\toprule
\underline{Input}    &   \multicolumn{2}{>{\columncolor[gray]{0.9}}c|}{Doc2Vec embeddings}   &    \multicolumn{3}{>{\columncolor[gray]{0.9}}c}{Word2vec embeddings}\\
{} &   \pst &               T2V &               KM & Spherical-KM &      GMM \\
\midrule
2  &                   &  0.6931* &                 &              &         \\
4  &           1.3863* &          &                 &              &         \\
5  &                   &          &         1.6094* &              &  1.6094 \\
6  &                   &          &         1.7917* &       1.7917 &  1.7917 \\
7  &           1.9458* &          &         1.9458* &       1.9458 &  1.9458 \\
10 &                   &          &         2.3025* &       2.3025 &  2.3025 \\
11 &                   &          &          2.3025 &      2.3978* &  2.3978 \\
12 &                   &          &         2.4848* &       2.4848 &  2.4848 \\
13 &           2.5648* &          &         2.5648* &       2.5648 &  2.5648 \\
15 &                   &          &         2.7082* &       2.7079 &  2.7079 \\
20 &  \textbf{2.9955}* &          &         2.9955* &       2.9442 &  2.9955 \\
22 &                   &          &         3.0908* &       3.0908 &  3.0632 \\
25 &                   &          &  \textbf{3.25}* &       3.2186 &  3.1666 \\
\bottomrule
\end{tabular}
\caption{Specificity values for Google News data set.}
\label{tab:gn_specif_tired}
\end{minipage}

\vspace{2mm}

\begin{minipage}[l]{0.99\columnwidth}
%
\caption{Dissimilarity values for Reuters data set.}
\label{tab:re_dissim_tired}
\end{minipage}
\qquad
\begin{minipage}[l]{0.99\columnwidth}
%
\caption{Dissimilarity values for Google News data set.}
\label{tab:gn_dissim_tired}
\end{minipage}

\raggedright
\small
\subsection{Computational Load}
The cosine similarity matrix $C$ grows quadratically with the number of documents to be processed.
Therefore the runtime order is $O((n*(n-1))/2)$.

In our Pytorch implementation, however, the matrix operation is highly optimized, and the computational time is reasonable.
For example, in our experiments, we used a GPU NVIDIA GeForce GTX 1080 Ti.
Even with Wiki20K, the largest data set we analyzed, 
training \pst\ took about 3 minutes, including the computation of two performance metrics, $C_V$ and NPMI.
Computing the two most time-consuming metrics,
training time rose to about 5 minutes.
Since \pst's training requires a grid search to optimize $\alpha$ and $\beta$, these computational times need to be multiplied for the number of $\alpha$ and $\beta$ combinations considered.

\end{table*}

\end{document}